\documentclass{article}

\usepackage{color}
\usepackage{PRIMEarxiv}

\usepackage[utf8]{inputenc} 
\usepackage[T1]{fontenc}    
\usepackage{hyperref}       
\usepackage{url}            
\usepackage{booktabs}       
\usepackage{amsfonts}       
\usepackage{nicefrac}       
\usepackage{microtype}      
\usepackage{lipsum}
\usepackage{fancyhdr}       
\usepackage{graphicx}       
\graphicspath{{media/}}     

\usepackage{amsmath}
\usepackage{amsfonts}
\usepackage{ amssymb }
\usepackage{dsfont}
\DeclareMathOperator*{\argmax}{argmax}
\DeclareMathOperator*{\argmin}{argmin}
\usepackage{mathtools}
\DeclarePairedDelimiter\ceil{\lceil}{\rceil}
\usepackage{hyperref}
\newcommand{\defines}{\triangleq}
\usepackage{natbib}
\setcitestyle{authoryear,open={(},close={)}}

\usepackage{relsize}
\usepackage{bm}

\title{What Makes a Good Explanation?: 
A Harmonized View of Properties of Explanations
}

\author{
  Varshini Subhash$^*$, Zixi Chen\thanks{Equal Contribution} \ , Marton Havasi, Weiwei Pan, Finale Doshi-Velez\\
  John A. Paulson School of Engineering and Applied Sciences \\
  Harvard University \\
  \texttt{\{zixichen, varshinisubhash\}@g.harvard.edu} \\
  \texttt{\{mhavasi, weiweipan\}@g.harvard.edu} \\
  \texttt{finale@seas.harvard.edu} \\
}

\begin{document}
\maketitle

\begin{abstract}

There have been many efforts that define ways to describe explanations---e.g. fidelity or compactness---but there is an unfortunately lack of standardization when it comes to defining properties of explanations. Different papers may use the same term to mean different quantities, and different terms to mean the same quantity. This lack of a standardized terminology of how we describe ML explanations prevents us from both rigorously comparing interpretable machine learning methods and identifying what explanation properties are needed in what contexts. 

In this work, we survey the mathematical formulations used to describe the properties of explanations in interpretable machine learning papers, synthesize them based on what they actually measure about the explanation, and describe the trade-offs between different formulations of these properties. In doing so, we enable more informed selection of task-appropriate formulations of explanation properties as well as standardization for future work in interpretable machine learning.
\end{abstract}


\section{Introduction}
Interest in interpretable machine learning\footnote{The properties used to describe explanations can apply both to inherently interpretable models as well as explanations derived from models.  Thus, in this paper, we will use the term interpretable machine learning synonymously with explainable AI.} has grown in recent years.  Regulatory bodies see interpretable machine learning as a method for auditing algorithms for safety, fairness, and other criteria---especially in high-stakes situations. Industry sees interpretable machine learning not only as a mechanism to provide oversight over their products, but also as a way to uncover insights on trends in their data and facilitate human+ML teaming.

It is well-accepted that different contexts will require different kinds of methods for interpretable machine learning. For example, the kind of explanation required to determine if an early cardiac arrest warning system is ready to be integrated into a care setting is very different from the type of explanation required for a loan applicant to help determine the actions they might need to take to make their application successful. Identifying which interpretability methods are best suited to which tasks remains a grand challenge in interpretable machine learning.

The \emph{computational properties} of an explanation may provide some clues towards whether an explanation will be useful in a particular setting. For example, consider an explanation that lists features a patient might change to reduce future cardiac risk and another explanation describing features a loan applicant might change to successfully get a loan.  In both cases, it is important that the list of features have no false positives---if the person acts on the listed features, the output should change. In both cases, it might also be important for the explanation to be reasonably short as the person may not have the time or inclination to parse through a complex description. In contrast, consider a scientist scrutinizing an explanation of a complex model that makes accurate predictions protein interactions in the context of discovery.  In this case, the scientist may find false positives---which they can weed out, based on domain knowledge---more acceptable than false negatives---areas of potential science left hidden. They may also be willing to spend much more time inspecting a longer explanation.

The examples above suggest that in many cases, we may be able to vet whether an explanation is likely to be useful in a particular context based on whether the explanation has the properties required for that context.  If one knows what properties are needed for what contexts, then one might be able to check for those properties computationally to identify promising interpretable machine learning methods prior to more expensive user studies with people.  In this way, properties can serve as an abstraction between the interpretable machine learning method and the context.

Unfortunately, while many works have defined properties, there is little current consensus around the terminology and formulation of interpretable machine learning properties. As a result, different works have used different terms for the same property. For example, one work might call a property robustness while another calls it stability or even fidelity. Different works have also mathematically formalized these properties differently: the expressions that one work uses to computationally evaluate compactness may be different than another. The current state of having multiple definitions in literature and a lack of consistent formulations makes it difficult to compare methods rigorously. It also becomes challenging to interpret what it truly means when one claims that a certain context needs a certain property or that a certain explanation has a certain property.

This paper reviews and synthesizes existing properties and their mathematical formulations in the interpretable machine learning literature. Our contributions are as follows:
\begin{enumerate}
    \item We collect together mathematical formulations describing the same properties but termed differently in different works, and standardize their notation.
    \item For each collection, we describe the substantive differences between the mathematical formulations and how those different ways of formalizing the property may be appropriate in different contexts. 
    \item We discuss the relationships and trade-offs between different properties and make precise how they may or may not be in tension with each other.
\end{enumerate}
In doing the above, our framework provides a much-needed systematic review and synthesis of a key part of the interpretable machine learning ecosystem. Our work serves as a reference for not only what are common properties that we might desire of interpretable machine learning methods, but also what considerations might go into making an abstract property mathematically precise. Our work serves as a reference for both what terms are used to describe properties as well as how to formalize them for future research in interpretable machine learning.

\section{Related Work}

The explosion of work in explainable and interpretable machine learning has prompted a number of review articles categorizing existing literature based on explainability methods, explanation properties, and methods for evaluation. To our knowledge, there has not been prior work that focuses on the \emph{mathematical definitions} of explanation properties.  Understanding the ways in which the abstract notions of properties have and can be made precise is crucial for cleaning up our current confusing state of multiple definitions, enabling future work to compare interpretable machine learning methods precisely, and ultimately understanding what information is needed in what contexts.

\paragraph{Reviews of Explanation Methods:}
There are several works which review and organize explanation methods proposed in literature. \citet{Arya2019} present users with a taxonomy to help them navigate all available explanation methods and arrive at the one most suited to their task. When discussing explanation evaluation, they consider two properties---\textit{faithfulness} and \textit{monotonicity}, which do not adequately cover the landscape of explanation properties. \citet{Carvalho2019} and \citet{Zhou2021} provide a detailed review of the field of machine learning interpretability and classify explanations by the methods used, the results they produce and their scope. With respect to evaluation, they list various explanation properties conceptually but do not delve into their mathematical definitions. \citet{Zhang2021} also divide the landscape based on explanation method and scope. \citet{Marcinkevics2020} provide a similar non-exhaustive categorization of explanation methods with a brief discussion on explanation properties. Similarly, \citet{Arrieta2020}, \citet{Adadi2018}, \citet{Mi2020}, \citet{Murdoch2019}, \citet{Linardatos2021}, \citet{Li2022}, \citet{Ras2018}, \citet{Fan2021}, \citet{Schwalbe2023} and \citet{Notovich2023} categorize explanation methods in detail and provide a high-level discussion of properties they must satisfy. \citet{Gilpin2018} classify explanation methods and find three broad categories--explanations which are simpler proxies of the original model, explanations which explain the model's internal data representation and self-explaining models. \citet{Guidotti2018} review explanation methods in addition to mapping them to the type of underlying model being explained, thus aiding user choice. All these works focus on the explanation types and methods and provide cursory discussions around their properties. In contrast, we provide a detailed categorization of the mathematical definitions of explanation properties across literature thus far. 

\paragraph{Reviews of Explanation Properties:}
When it comes to evaluating explanation quality, the terms \emph{property} and \emph{evaluation metric} are often used synonymously in different works. We prefer the term \emph{property} to avoid conflation with the ultimate downstream evaluation metric---how well an explanation aids a user in performing their task. 

\citet{Zhou2021} review explainable literature, select the concepts most relevant to explanation quality and identify recurrent properties which they deem as important. Further, they map existing explanation methods to these properties and highlight gaps where the method does not satisfy a particular property. Our work improves upon this list of properties by unifying other instances of them across literature, and describing them in a mathematically precise manner. \citet{Schwalbe2023} and \citet{Vilone2021} also provide a qualitative review of explanation properties, in contrast to our mathematically motivated review. The work that comes closest to ours is by \citet{Nauta2022}, who identify 12 overarching properties by systematically surveying explanation evaluation across 300 XAI papers published over the past 7 years. However, our work differs from theirs in some important ways. In contrast to their qualitative discussion of these properties, we consider their precise mathematical definitions, which allows us to unify the proposed 12 properties to 4.  They also exclude all properties which pertain to model fairness, safety and privacy---which we argue are aspects crucial towards the discussion of explanation quality, and are hence included in the scope of our work.  

There are other works that survey properties and identify gaps by comparing against requirements from other domains. \citet{Sovrano2021} offer a mapping between explanation properties in present literature and legally mandated properties for high-quality explanations, as prescribed by the EU Artificial Intelligence Act. We differ by restricting our focus to the harmonization of mathematical properties introduced specifically in the interpretable machine learning literature.

\paragraph{Reviews of Explainability for Application Domains:} 
Explainability has also been reviewed from the perspective of the needs of a  specific application domain. \citet{Nunes2017} provide a systematic review of explanations for decision-support and recommender systems. They categorize based on the explanation content, presentation, and human-centric properties. 
\citet{Abdullah2021} survey explainable literature pertaining to healthcare, where transparency and interpretability is paramount. They categorize explanation methods, tie them to use cases in healthcare and qualitatively discuss explanation properties. \citet{Markus2021} also categorize explanation methods, list properties they deem necessary from the healthcare context, and evaluate the extent to which these properties are satisfied by existing methods. \citet{Chromik2020} and \citet{Lim2019} adopt the human-computer interaction (HCI) lens, where they review design decisions for user studies targeted at the usage and evaluation of explanations. \citet{Theissler2022} review explanation methods from the perspective of time-series classification while \citet{Yuan2023} do so for graph neural networks, with a brief discussion on explanation properties. In contrast, we focus on the mathematical definitions of computational (not human-centric) properties and keep our selection domain-agnostic.

\paragraph{Other Taxonomies:} 
There are other XAI taxonomies which do not fit into the aforementioned categories but are aimed at unifying explainable literature. \citet{Graziani2023} review machine learning interpretability and its usage across various application domains and unify previously scattered terminology and concepts in the field. \citet{Zytek2022} present a taxonomy on explainable features as opposed to explanations, where they consider users across various domains and the human-centric properties to be satisfied for features to be explainable. \citet{Speith2022} and \citet{Sogaard2022} offer a taxonomy of explainability taxonomies, highlight the challenges and weaknesses in their construction and offer solutions towards arriving at a consistent view of the explainability landscape. 

\section{Notation and Terminology}
\newcommand{\x}{\mathbf{x}}
\newcommand{\E}{E}
There are many notation and terminology conventions in the literature.  Below, we define the notation and terminology we will use in this work.  All equations from other works will be converted into this notation for ease of comparison; however, we will also reference the original equations to allow the reader to refer to the source.

Throughout the work, we look at predictive models $f$ that yield a prediction $\hat{y}=f(\x)$  for some $K$-dimensional input $\x \in \mathcal{X}$, where $\mathcal{X}$ is the training dataset. We use $x^{(k)}$ to refer to the $k^{\texttt{th}}$ dimension of $x$. For the model $f$, we denote its prediction as $\hat{y}_f$, which may be discrete or continuous depending on whether the task is classification or regression. 

We use $\E(f)$ to represent the explanation; we define the \emph{explanation} as the information provided from the model to the user. In the case of inherently interpretable machine learning models, $\E(f) = f$.  If the explanation depends also on the input, we will use the notation $\E(f,\x)$.  Certain evaluations of explanations also require a baseline or reference input value. We denote $\x_0$ as this reference value. Some also take into account the ground truth value at an input $\x$, and we denote $y$ as the respective ground truth. 

Explanations in the interpretable machine learning literature tend to fall into three main categories: function-based explanations, feature-attribution-based explanations, example-based explanations:

\paragraph{Function-based Explanations:}
We use the term function-based explanations for models or structures that are inherently interpretable and allow one to produce an output or reasoning given an input. Typically, this means that the explanation will produce a prediction $\hat{y}_\E$ that can be compared to the model prediction $\hat{y}_f$. Examples are: a local surrogate explanation that uses an interpretable model (e.g. sparse linear model, decision tree, etc.) to approximate the target model locally around an individual prediction; a decision set that approximates the target model's behavior globally with nested if-else rules; a self-explaining neural network that has an interpretable linear form but with the coefficients being neural networks and depending on the input; a concept bottleneck model that maps features to interpretable concepts and then uses the concepts to predict the output. 

\paragraph{Feature Attribution Explanations:}
Feature attribution explanations identify the input features that had a key role in producing the output (perhaps ordered or weighted somehow). All feature attribution explanations can be described by a length-$K$ vector, in which each entry $E(f, \x)_k$ is the attribution score for each input dimension $k = 1 \dots K$ at observation $\x$ for model $f$. Different feature attribution methods use different ranges of values for $E(f, \x)_k$: some may assign both positive and negative weights; others may only assign non-negative weights. The \emph{ranking} of features $\text{Rank}_{\E}(f,\x)$ given the attribution weights $E(f, \x)_k$ is the ordering of the features from largest to smallest weight.

\renewcommand{\S}{\mathcal{S}}
Sparse feature attribution methods attempt to identify a subset of relevant features from the full set. We use $\S\subseteq \{1\dots K\}$ to represent the retained subset of features. We use $\S_\x$ to denote the set of retained features for input $\x$ given to the explanation $\E(f,\x)$. 
We use $\x_\S$ to denote a version of the input $\x$ for which the values of the features in $\S$ are retained and the values of the features not in $\S$ are reverted to baseline values $\x_0$: $\x_\S^{(k)}=\x^{(k)}$ if  \, $k \in \S$; otherwise $\x_\S^{(k)} = \x_0^{(k)}$. Similarly, we use $\S^c$ to denote the complement of $\S$ and $\x_{\S^c}$ to denote the input $\x$ with features in $\S^c$ retained and features in $\S$ reset to the reference value $\x_0$, i.e. $\x_{S^c}^{(k)} = \x_0^{(k)}$ if $k \in \S$; otherwise $\x_{S^c}^{(k)} = \x^{(k)}$. 

Unlike function-based explanations, a list of feature attributions does not, in itself, provide a way to predict an output given an input. However, one common approach to making predictions given a feature attribution-based explanation $E(f, \x)$ is to compute the prediction $f(\x_\S)$.

\paragraph{Example-based Explanations:} Example-based methods select a subset of representative samples from the dataset to explain model behavior or the underlying distribution.



\section{Overview of Properties of Explanations}

There are many properties of explanations that have been described in the literature, ranging from fidelity to robustness to privacy. For this work, we focus on computational properties, that is, properties of explanations that can be described by a mathematical formula.  Our goal is harmonize and synthesize the many different mathematical formulations used in the existing literature.  To find works which define computational properties, we first surveyed all papers at ICML and NeurIPS going back five years to identify papers relating to explainability.  From those papers, we identified further citations (in their literature review) and also created a list of key terms used to describe the computational properties of explanations.  We searched for each of those key terms on Google Scholar.  Finally, we collated the property definitions across all the papers from our initial survey of recent conference papers, older conference papers, and Google Scholar results associated with those key terms. 

\begin{figure}[htp!]
\centerline{\includegraphics[width=1.1\columnwidth]{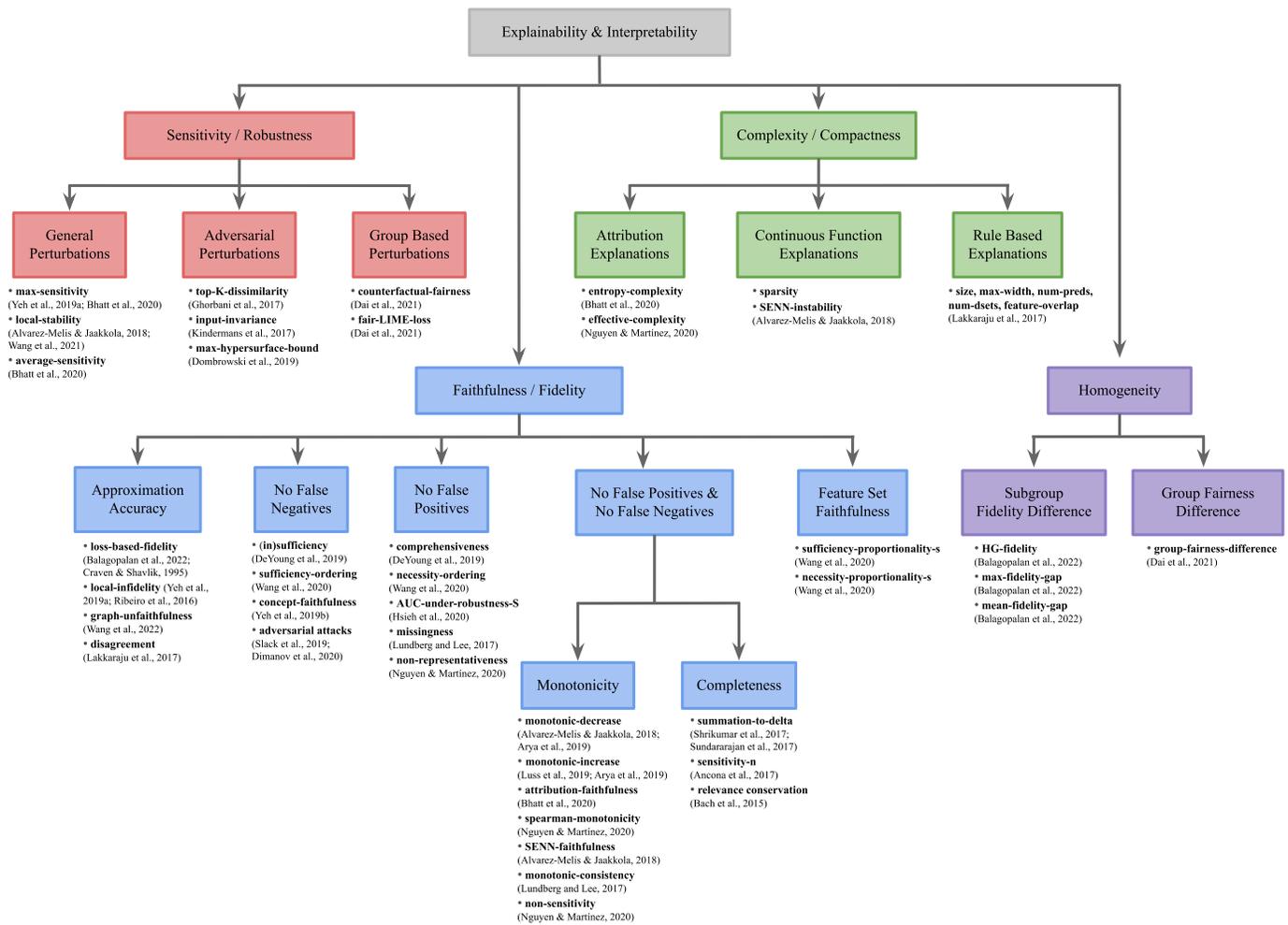}}
\caption{Property Synthesis Framework}
\label{fig:property-synthesis}
\end{figure}

From these works, we found that almost all the computational properties of explanations fell into one of the four broad categories (visualized in Figure~\ref{fig:property-synthesis}): Robustness, fidelity, compactness, and homogeneity.  Below, we briefly define the abstract notion of each of these properties, and then in the next sections we synthesize the varying mathematical formulations for each of these properties, in particular with an eye to how different mathematical formulations may be important in different settings.  We also briefly summarize other computational properties in Section~\ref{sec:discussion}. We emphasize again that these properties apply to both explanations derived from models as well as explanations that are the model (the inherently interpretable case).  For example, we can talk about the effort required to process an explanation---its compactness---in either case.

\paragraph{Robustness / Sensitivity:} \emph{Robustness} or \emph{sensitivity} measures how much the explanation is prone to change when the input $\x$ is changed (especially infinitesimally or imperceptibly).  Robustness is important because users often assume that explanations for one input will apply to (what they perceive as) other similar inputs.  If they assume incorrectly, they could make an incorrect use of an explanation.  To some extent, the robustness of the explanation is tied to the curvature of the function $f$; however, it is also possible for explanation methods to output explanations that change more rapidly than the underlying function \cite{Yeh2019OnT}.

\paragraph{Faithfulness / Fidelity:} \emph{Faithfulness} or \emph{fidelity} evaluates the explanation's capability to capture the true underlying decision-making process of the model.  If an explanation does not reveal the model's true decision-making process, then a user could make incorrect inferences based on the incorrect (or partial) information.  These incorrect inferences could lead to incorrect decisions by the users and thus unfavorable outcomes, especially in high-risk scenarios.

\paragraph{Complexity / Compactness:} \emph{Complexity} or \emph{compactness} describes the cognitive effort that users would have to exert to understand the explanation.  We define complexity with respect to cognitive effort, rather than simply explanation size, as explanations that require less cognitive effort to be used are likely to be preferred---and more accepted and used---by users.  That said, we will see that computational definitions of complexity do come down to various notions of explanation size. There are often tensions between accurately capturing the model's decision-making process (faithfulness) and the complexity of the explanation.

\paragraph{Homogeneity:} \emph{Homogeneity} refers to whether some property---such as fidelity---has the same value across explanations for different inputs.  In particular, we are usually interested in whether explanations (for the same model) have the same properties across different subgroups.  In this sense, homogeneity can be viewed as the explanation's robustness to input perturbations to subgroup membership. When these subgroups differ by a sensitive demographic attribute, homogeneity becomes related to fairness: it can be problematic if explanations are more accurate for certain subgroups than others.  

\section{Robustness and Sensitivity}


The first property we examine is \emph{robustness}, also often referred-to as \emph{sensitivity}. For local interpretability methods (i.e. methods that explain the prediction for a given $\x$), sensitivity measures the similarity of explanations under changes to the input point $\x$. For global interpretability methods which are concerned with explaining the entire model, the explanation does not change with respect to the input $\x$. Thus, issues around robustness do not apply. However, a global explanation may still have issues around complexity, which we discuss later (Section~\ref{sec: complexity}).

In general, users expect explanations to be stable under minor changes to the input point $\x$. If a user sees an explanation for $\x$ and incorrectly presumes it also applies to a slightly changed input $\x'$, that could result in them making an incorrect decision. In this way, a lack of robustness can also allow users to be manipulated via adversarial attacks that perturb the input imperceptibly and change the explanation to one that the adversary wishes shown. More broadly, \citet{Yeh2019OnT, Ghorbani2017NNFragile} demonstrate that robustness (appropriately) increases user trust in the explanation: users expect and trust robust explanations.
Explanation consistency also alludes to a lack of arbitrariness in the model's decision-making, improves the user's ability to predict the model's strengths and weaknesses and minimizes risk during deployment. 

\paragraph{Explanation Sensitivity vs. Model Sensitivity}
Before continuing regarding methods to measure \textit{explanation sensitivity}, we emphasize that \textit{model sensitivity} and \textit{explanation sensitivity} are two related but distinct concepts. \textit{Model sensitivity} is how sensitive the output of the model $\hat{y} = f(\x)$ is to the input $\x$.  \textit{Explanation sensitivity} is how sensitive the explanation $E(f,\x)$ is to the input $\x$. They are related in that a faithful explanation of a quickly-changing function (highly sensitive model) may also need to be quickly-changing.  However, it is possible for an explanation to have greater sensitivity than the underlying model, or a model to be more sensitive than the explanation.

The relationship between model sensitivity and explanation sensitivity, and their effect on explanation faithfulness, was first studied theoretically \cite{Yeh2019OnT}. \citet{Tan2023RobustForFree} then proved that robust models are guaranteed to produce more robust explanations than their non-robust counterparts. They seek to understand the effect of model non-robustness on explanation robustness and what happens when the explanation is sensitive despite a robust model. The former is less interesting than the latter, and can also be viewed as an inevitable trade-off between explanation faithfulness and explanation sensitivity -- if a model is too sensitive and the explanation is explaining the model faithfully, then we lose explanation robustness in the bargain. In order to handle such a scenario in practice, various approaches have been proposed. Some ensure robust explanations via \textit{model smoothing}, i.e. by learning smoother functions \citet{Smilkov2017, ross2018, Wang2020smoothedgeometry, Xue2023Stability, Tan2023RobustForFree}. Some perform \textit{model retraining} by including adversarial examples in the training data and empirically show higher explanation robustness \citet{Tsipras2019RobustnessOdds}. We discuss the inevitable robustness-faithfulness trade-off in such situations in Section~\ref{subsec:faithfulness-sens}.

That said, it is also possible for the explanation to be more sensitive than the underlying function. For example, a popular way to explain a function is via its gradients: $E(f,x) = \nabla_f(x)$, \citet{baehrens2009, simonyan2014deep, springenberg2015striving, Bach2015OnPE, AxiomaticAttribution, Selvaraju2016}. However, the raw gradients of a model tend to be noisy \citet{Samek2021Review, Smilkov2017, Ghorbani2017NNFragile}, leading to higher explanation sensitivity. In fact, \citet{Dombrowski2019} show that if we consider the hypersurface of inputs which generate constant model output, this hypersurface appears to have high curvature, i.e. the model's gradient is highly sensitive to perturbations. This motivates \textit{model smoothing} for obtaining robust explanations. Another approach is \textit{explanation smoothing}, which modifies the explanation to be an average of multiple explanations in the surrounding region, and can result in explanations that are \textit{both} more stable and more faithful to the underlying model $f$, \citet{Yeh2019OnT, Smilkov2017, Omeiza2019SmoothGradCam, Wang2020ScoreCAM, Naidu2020, Raulf2021, Gan2022MeTFA, Ajalloeian2022SmoothedExp, Rieger2020SimpleDefense}. Various other techniques \citet{Jha2021, Zeng2023, Lim2021, Dombrowski2022RobustExp} have also been explored to achieve more robust yet faithful explanations. 

\paragraph{Types of Robustness Definitions}
As sensitivity is most commonly defined as how the explanation changes with respect to changes in the input, a key decision is to determine what kinds of input perturbations are appropriate for the situation. Common choices of perturbations include considering how explanations change within a sphere of a given radius surrounding the input $\x$. Group perturbations are performed by perturbing group membership values of the input $\x$. \citet{Dandl2023Hyperbox} use a \textit{hyperbox} as the perturbation region, to perform such group perturbations without affecting model output. In this case, the set of allowable input perturbations in the hyperbox serves as an explanation itself. Within the region defined by the allowable set of input perturbations, one can be interested in some notion of average change in explanations or the worst case change to the input (an adversarial input perturbation). 

In general, when measuring sensitivity of a function (an explanation in our case), one is concerned with how rapidly the function changes with changes to the input. When it comes to defining mathematical metrics to measure sensitivity, some common themes that can be leveraged include Lipschitz continuity, function smoothing via averaging and regularization -- all of which come with useful robustness properties. In all the sensitivity metrics presented, we see the above ideas motivating each formulation, with task-dependent variations.  Specifically, in the remainder of the section, we describe sensitivity properties grouped by: 

\begin{itemize}
\item Sensitivity via \emph{general input perturbations} measures the change in explanation as a function of change in input, where this change in input is described via local perturbations of a given radius (under some distance metric). Different mathematical formulations output the maximum change in explanation which this spherical region, the average change in explanation, or the maximum ratio of explanation change to perturbation size. We describe these in Section~\ref{subsec:genperturb}.

\item Sensitivity via \emph{adversarial input perturbations} is a special case where the perturbations are intended to cause \textit{specific} change in explanation while the input is perturbed imperceptibly (e.g. stay within some radius of the original input). This imperceptible perturbation can have one of two outcomes. First, the model is robust to the perturbation, but the explanation isn't and misleads the user. Second, the model is sensitive to it and hence the explanation, which explains the sensitive model, is also sensitive to it. We consider both cases and describe each in Section~\ref{subsec:advperturb}. While most of the current applications of adversarial perturbations have been to images, the general definition can be applied to other use cases.



\item Sensitivity via \emph{group-based input perturbations} measures the change in an explanation by perturbing not the entire input, but a single attribute or subset of attributes in the input. This induces a perturbation region defined by a hyperbox \citet{Dandl2023Hyperbox}, where each attribute's perturbation is performed along an axis. A special variant of a group-based perturbation is a \textit{protected group-based perturbation}, where we change only a protected attribute in the input---e.g. gender or race. In this case, the perturbation is along a single axis: the protected attribute. If an explanation is not sensitive to a protected attribute and faithfully explains the underlying model, i.e. it does not respond to a change in a protected attribute unless the model does---then it can reveal if a model is biased or unfair.
We describe these in Section~\ref{subsec:groupperturb}.

\item Sensitivity via \emph{model perturbations} measures the change in an explanation by perturbing the model instead of the input. Changing the model is bound to yield a different explanation, but measuring the exact impact of the former on the latter is useful. For instance, we can help mitigate explanation manipulation caused by adversarial model manipulation. Some findings in this space are even favorable towards robust explanations--perturbing the model parameters and performing explanation smoothing has been found to yield less sensitive explanations. We describe these in Section~\ref{subsec:modelparameterperturb}.

\item Sensitivity via \emph{explanation hyperparameter perturbations} refers to the sensitivity of the explanation to its own generation process. Sometimes, the explanation generation process involves hyperparameters and ideally, an explanation's output should be agnostic to changes to these hyperparameters. This has been shown to not be the case especially with feature attribution explanations for images. We describe these in Section~\ref{subsec:exp_hyperparams_sens}.

\end{itemize}

We expand on each below.

\subsection{Sensitivity to General Input Perturbations} \label{subsec:genperturb}
General perturbations usually refer to random perturbations within a specific region around the original input. Common choices in literature include spherical regions (according to some distance metric); when the distance metric only considers a single attribute, the input region becomes a cylinder. However, a general definition of sensitivity could use geometric regions of different shapes, depending on the application. When discussing sensitivity to general input perturbations, we only consider the case where the model is robust to the perturbation and the explanation is not.

The first metric of sensitivity via general perturbations, proposed by \citet{Yeh2019OnT} (also discussed by \citet{Bhatt2020}, \citet{Singh2019AttributionalRobustness}), defines \textbf{max-sensitivity} as the maximum change in the explanation $\E$ under a small perturbation. In its most general form, the perturbation is defined within a sphere of radius $r$ under $p$-norm $\ell_p$ around the input $\x$ and the change in explanation can be measured by any appropriate distance metric $\ell$. 

\begin{flalign}\label{maxsensgeneral}
&& \text{max-sensitivity}(\E,f,\x,r) \defines \max_{\ell_p(\x'-\x) \leq r} \ell(\E(f,\x'), \E(f,\x)) &&  \triangleright\,\text{Generalized definition 3.1 in \citeauthor{Yeh2019OnT}}
\end{flalign}

\citet{Yeh2019OnT} instantiate the metric using the $\ell_2$ norm as follows:

\begin{flalign}\label{maxsens}
&& \text{max-sensitivity}(\E,f,\x,r) \defines \max_{\left\Vert \x'-\x \right\Vert \leq r} \left\Vert \E(f,\x') - \E(f,\x) \right\Vert &&  \triangleright\,\text{Definition 3.1 in \citeauthor{Yeh2019OnT}}
\end{flalign}

While bounding max-sensitivity bounds the change in explanation within a small region, it does not require continuity. That is, the explanation may change abruptly within the region. In contrast, \citet{AlvarezMelis2018} (also discussed in \citet{Yeh2019OnT}) propose an alternative metric, called \textbf{local-stability}, that measures the ratio of the maximum change in explanation within the perturbation region to the size of the perturbation region (akin to Lipschitz continuity). Again, the change in the explanation $\E$ and the distance from $\x$ are both measured by the $\ell_2$ norm:
\begin{flalign}\label{localstab}
&&\text{local-stability}(\E,f,\x,r) \defines \max_{\left\Vert \x'-\x \right\Vert \leq r} \frac{\left\Vert \E(f, \x') - \E(f, \x) \right\Vert}{\left\Vert \x' - \x \right\Vert} && \triangleright\,\text{Equation 5 in \citeauthor{AlvarezMelis2018}}
\end{flalign}

\citet{normalattr, Wang2020smoothedgeometry, Agarwal2021Unification, Tan2023RobustForFree} -- making a more direct use of Lipschitz continuity -- refer to robustness as a constrained version of \textbf{local-stability} (Equation~\ref{localstab}), wherein an explanation is \textbf{lipschitz-locally-stable} if the Lipschitz continuity holds locally, for a given threshold $\lambda$: 

\begin{flalign}\label{localstab}
&& \text{local-stability} (\E,f,\x,r) <= \lambda && \triangleright\,\text{Definition 6 in \citeauthor{Wang2020smoothedgeometry}}
\end{flalign}

Similarly, an explanation is \textbf{lipschitz-globally-stable} for all $\x' \in \mathbb{R}^{K}$, a threshold $\lambda$ and $p$-norm $\ell_p$ if: 

\begin{flalign}\label{localstab}
&& \ell_p(\E(f, \x') - \E(f, \x)) <= \lambda \ell_p(\x'-\x) && \triangleright\,\text{Definition 6 in \citeauthor{Wang2020smoothedgeometry}}
\end{flalign}

However, \citet{AlvarezMelis2018} posit that enforcing this global condition may be undesirable in practice, in terms of trade-offs with fidelity.

Unlike \textbf{max-sensitivity}, bounding \textbf{local-stability} ensures that the perturbed input explanations become sufficiently similar to the explanation at input $\x$, as we approach $\x$.
In fact, as $r \rightarrow 0$, \textbf{local-stability} converges to the $\ell_2$ norm of the gradient of $\E(f, \x)$ at $\x$, so it can be easily approximated for small $r$ (assuming that $\E$ is differentiable). 

However, the disadvantage of bounding \textbf{local-stability}, aka \textbf{lipschitz-local-stability}, to a threshold $\lambda$ is that explanations become insensitive to all small input perturbations, including the ones that change the model prediction. Thus in practice, we always want to maximize robustness subject to a faithfulness constraint -- a property we discuss in Section~\ref{sec:faith}.

With both \textbf{max-sensitivity} and \textbf{local-stability}, it is difficult to compare sensitivity across different explanation methods---or ultimately use them for computing these properties---because each explanation may have a different range or magnitude. To address this, \citet{agarwal2022rethinking} introduce \textbf{relative-stability} which considers a percentage change in the explanation. Let $\ell_p$ be the $p$-norm and $\epsilon > 0$ be the minimum threshold used to prevent division by zero when $\x = \x'$.
\begin{flalign}\label{relstab}
&&\text{relative-stability}(\E,f,\x,r) \defines \max_{\left\Vert \x'-\x \right\Vert \leq r} \frac{\ell_p(\frac{\E(f, \x) - \E(f, \x')}{\E(f, \x)})}{\max ( \ell_p(\frac{\x - \x'}{\x}), \epsilon)} && \triangleright\,\text{Equation 2 in \citeauthor{agarwal2022rethinking}}
\end{flalign}
 
All formulations so far consider the change in the explanation with respect to a change in the input (according to some distance metric). In some cases, that appropriate distance metric may be the distance between the intermediate representations $R_\x$ and $R_{\x'}$ of nearby points $\x$ and $\x'$ in the model. That is, even if the inputs $\x, \x'$ and outputs $\hat{y}, \hat{y}'$ are similar, different neurons are being activated. This notion of distance between representations may be appropriate when an adversarially crafted $\x'$ appears to be imperceptibly similar to $\x$ to a human, but the model generates wildly different internal representations for both. In such cases, the perturbation region is more accurately captured by how far apart $\x$ and $\x'$ are in high-dimensional space via their representations. 
\citet{agarwal2022rethinking} call this metric \textbf{representation-stability}: 
\begin{flalign}\label{reprstab}
&&\text{representation-stability}(\E,f,\x,r) \defines \max_{\left\Vert \x'-\x \right\Vert \leq r} \frac{ \ell_p(\frac{\E(f, \x) - \E(f, \x')}{\E(f, \x)})}{\max ( \ell_p(\frac{R_{\x} - R_{\x'}}{R_{\x}}), \epsilon)} && \triangleright\,\text{Equations 3 and 5 in \citeauthor{agarwal2022rethinking}}
\end{flalign}
where the $\epsilon > 0$ is used to prevent division by zero. One drawback of this approach is that it assumes white-box access to the model. If the user has black-box model access, we can instead use final model outputs like logits instead of hidden representations. 

Finally, all of the mathematical formulations of robustness above contain a $\max$ over the difference between explanations in some perturbation region.  If the explanation varies smoothly over this region except for at a few isolated points, these mathematical formulations will be determined by those few outliers and not be representative of the overall smoothness of the explanation method. To address this, \citet{Bhatt2020} consider the average change in the explanation within the perturbation region.  Their mathematical formulation, \textbf{average-sensitivity}, takes the expectation of the difference between the explanation at the fixed input $\x$ and random inputs $\x' \sim p(\x)$:
\begin{flalign}\label{avgsensb}
&&\text{average-sensitivity}(E,f,\x,p) \defines \int_{\x' \in \mathbb{R}^K} \ell(\E(f, \x), \E(f, \x')) p(\x') \ \mathrm{d}\x' && \triangleright\,\text{Definition 2 in \citeauthor{Bhatt2020}}
\end{flalign}
In the simplest formulation, $p(\x)$ is uniformly distributed in a sphere with radius $r$ around $\x$: $p=U(\{\x'|\left\Vert \x'-\x \right\Vert \leq r\})$, though any general distance metric $\ell(\x', \x)$ can be used. 
We denote the change in the explanation with a general distance metric $\ell$. An additional benefit of \textbf{average-sensitivity} is that one can obtain an unbiased estimate by drawing Monte-Carlo samples from $p(\x)$, while the other formulations in this section require some way to compute the maximum difference. It is also relatively easy to optimize for post-hoc, via explanation averaging or explanation aggregation.

\subsection{Sensitivity to Adversarial Input Perturbations} \label{subsec:advperturb}
While general perturbations usually refer to all perturbations in a geometric region surrounding the original input, the goal of an adversarial perturbation is usually to change the explanation in a certain way without changing the input $\x$ significantly. Adversarial perturbations resemble finding the worst-case perturbation, as was the case with maximum sensitivity described in Equation~\ref{maxsens}, but with this distinct goal of manipulation. There are two potential scenarios for why the perturbation changes the explanation---a robust model produces an unchanged output $\hat{y}$, but its explanation is sensitive to the perturbation, or a sensitive model responds to the perturbation with an incorrect output $\hat{y}$ and this yields a sensitive explanation. Below we focus on the former case, because if the model is also sensitive, then we have a larger problem. (And in Appendix~\ref{app:robust_exp_via_model}, we describe model-smoothing techniques that have been used to address the latter.)


An adversarial perturbation can affect a robust model minimally and still produce a sensitive explanation. For example, \citet{Ghorbani2017NNFragile} highlight that in the medical domain, images could be adversarially perturbed to offer misleading causal insights---eg. a pixel-based explanation could divert attention from the location of a malignant tumor. Such cases are especially hard to detect because changes to the input image are not visible to the human eye and the model's predicted labels remain unchanged. \citet{Salahuddin2022} review explanations for medical image analysis and highlight robustness as an essential property towards trustworthy explanations in the clinical setting.

In addition to general perturbations, the formulation of \textbf{max-sensitivity} (Equation~\ref{maxsensgeneral}) naturally lends itself to the context of adversarial perturbations. An adversary is interested in finding the worst-case perturbation within a perturbation region which maximizes an explanation's sensitivity, but retains model output. \citet{Singh2019AttributionalRobustness}, \citet{Wang2020smoothedgeometry}, \citet{Ghorbani2017NNFragile} employ \textbf{max-sensitivity} specifically for the application of image-based adversarial perturbations, where the similarity metric $\ell$ can be the Spearman and Kendall rank correlation coefficients ~\ref{spearman_sens}, ~\ref{kendall_sens}, top-$k$-intersection ~\ref{topkintersection}, or cosine-dissimilarity ~\ref{cosine_dissim_sens}. 

While much of the literature on adversarial perturbations is in the context of images \citet{Ghorbani2017NNFragile, Wang2020smoothedgeometry, Singh2019AttributionalRobustness, Kindermans2017, Dombrowski2019}, the concept applies to any situation in which someone may wish to manipulate an explanation. Specifically, \citet{Ghorbani2017NNFragile} consider adversarial perturbations to manipulate feature importance (pixel-based) explanations for images. Their general approach involves using an iterative process to find an $\x'$ in a perturbation region whose explanation has the greatest distance from $E(f, \x)$. The formulations below describe how they measure this explanation distance, i.e. sensitivity.

Let $\S_{\x} = \{1\dots k\}$ be the subset of features which have the top $k$ (highest) importance scores given by $\E(f,\x)$. Let each importance score given by $\E(f,\x)$, for a feature in $\S_{\x}$ be $E(f,\x)_i$, where $i \in \S_{\x}$. For a perturbed input $\x'$, let the importance scores given by $\E(f,\x')$ for a feature in $\S_{\x}$ be $E(f,\x')_i$. The mathematical formulation for \textbf{top-\textit{k}-dissimilarity} can then be defined using the sum of those $k$ importance scores given a perturbed input $\x'$, which were highest scoring given the input $\x$. For a worst-case adversarial perturbation, this sum should be as low as possible, which when negated gives a `higher-is-better' dissimilarity value. In other words, this formulation can be viewed as trying to make the features ranked most highly for input $\x$ as unimportant as possible for the perturbed input $\x'$. This is the same as finding a perturbed explanation that is farthest from the unperturbed explanation as in \textbf{max-sensitivity} (Equation~\ref{maxsens}), with the difference that the region of perturbation is defined as the top $k$ most important features for an input $\x$ and not all features.

\begin{flalign}\label{topkdissimilarity}
&&\text{top-\textit{k}-dissimilarity}(\E, f, \x, \x') \defines \max_{\left\Vert \x'-\x \right\Vert \leq r}  \left[-\sum_{i \in \S_{\x}} E(f,\x')_i\right] && \triangleright\,\text{Algorithm 1 in \citeauthor{Ghorbani2017NNFragile}}
\end{flalign}

A variant of the formulation above is \textbf{targeted-dissimilarity}, wherein the user has a predetermined region (of unimportance) in the input image $\x$ defined by the set of features $\S_{\x}$. Instead of picking the top $k$ highest scoring features, these features which are deemed unimportant or misleading can be picked, thus allowing the user the flexibility to \text{target} specific features. The higher the sum of importance scores for these misleading features, the better. Similar to the argument above, this formulation resembles \textbf{max-sensitivity} (Equation~\ref{maxsens}), where the region of perturbation is a user-defined set of unimportant features for an input $\x$.

\begin{flalign}\label{targeteddissimilarity}
&&\text{targeted-dissimilarity}(\E, f, \x, \x') \defines \max_{\left\Vert \x'-\x \right\Vert \leq r}  \left[\sum_{i \in \S_{\x}} E(f,\x')_i\right] && \triangleright\,\text{Algorithm 1 in \citeauthor{Ghorbani2017NNFragile}}
\end{flalign}

The third formulation is \textbf{center-of-mass-dissimilarity} which computes the dissimilarity between the centers of masses of the perturbed and unperturbed images. The \textbf{center-of-mass} $\mathcal{C}(\x)$ of an image $\x$ is defined as weighted sum of all pixel values of $\x$, where the weights are the corresponding feature importances given by $E(f,\x)$. The formulation computes the $\ell_2$ distance between the centers of masses. This formulation resembles \textbf{max-sensitivity} such that the region of perturbation is characterized by the center of mass of the input $\x$.

\begin{flalign}\label{centerofmassdissimilarity}
&&\text{center-of-mass-dissimilarity}(\E, f, \x, \x') \defines \max_{\left\Vert \x'-\x \right\Vert \leq r}  \left\Vert \mathcal{C}(\x) - \mathcal{C}(\x') \right\Vert && \triangleright\,\text{Algorithm 1 in \citeauthor{Ghorbani2017NNFragile}}
\end{flalign}


\citet{Kindermans2017} define \textbf{input-invariance} as the change in explanation for an input perturbation $\x' = \x + c$ where $c \in \mathbb{R}$. This perturbation does not impact model output, hence should ideally not impact the resultant explanation either. While the authors measure sensitivity qualitatively through a visual comparison of pixel-based explanations, a constrained version of \textbf{max-sensitivity} with $\ell_1$ and $\ell_2$ norms could capture this quantitatively. In the ideal scenario where $E(f,\x)$ and $E(f,\x')$ are perfectly identical, \textbf{input-invariance} as defined below, would be zero and the higher its value, the more the explanation sensitivity. For $p=\{1, 2\}$, consider:

\begin{flalign}\label{input-invariance}
&&\text{input-invariance}(\E, f, \x, \x') \defines \max_{\left\Vert \x'-\x \right\Vert = c} \ell_p(\E(f,\x'), \E(f,\x)) && \triangleright\,\text{Section 3.1 in \citeauthor{Kindermans2017}}
\end{flalign}



\citet{Dombrowski2019} extend the idea of using a targeted subset of unimportant features as seen with \textbf{targeted-dissimilarity}, and perform regularization with an adversarial target image. Let $\x_{\text{target}}$ be an arbitrary target image chosen by an adversary, with features (pixels) that are irrelevant to the model output and would be misleading if used as an explanation. The formulation \textbf{targeted-regularizing-loss} penalizes a large difference in model outputs for inputs $\x$ and $\x'$, thus encouraging similar model outputs. At the same time, it penalizes a large difference in the perturbed explanation $\E(f,\x')$ and the target image $\x_{\text{target}}$.

\begin{flalign}\label{target}
&&\text{targeted-regularizing-loss}(\E, f, \x') \defines \left\Vert \E(f,\x') - \x_{\text{target}} \right\Vert ^ {2} + \lambda \cdot \left\Vert f(\x') - f(\x) \right\Vert ^ {2} && \triangleright\,\text{Equation 4 in \citeauthor{Dombrowski2019}}
\end{flalign}

Further, \citet{Dombrowski2019} argue that the hypersurface of inputs generating constant model output $\mathcal{X} = \{\x \in \mathbb{R}^{d} | f(\x) = c\}$ for a constant $c$, is likely to have high curvature. By reducing this curvature via \textit{model smoothing}, one can experimentally obtain more robust explanations. They propose \textbf{$\bm{\beta}$-smoothing} where they replace the model's ReLU activation with a Softplus activation. 
A related approach that tackles explanation sensitivity to adversarial attacks given a robust model is \textit{explanation smoothing}. This involves averaging multiple explanations and has been shown to empirically produce more robust explanations \citet{Rieger2020SimpleDefense}.


\subsection{Sensitivity to Group Based Input Perturbations}\label{subsec:groupperturb}
Group-based perturbations consider the very specific perturbation region induced (i.e. a hyperbox \citet{Dandl2023Hyperbox}) by perturbing a single input attribute or a subset of input attributes, in a $K$-dimensional input $\x$. All the remaining input features remain the same. A special case of this is perturbing a single protected attribute, which we call a \textit{protected group based perturbation}. This case is interesting due to its implications on fairness. For instance, suppose we have gender as a protected feature with group values contained in the set \{\emph{male}, \emph{female}, \emph{non-binary}, ... \}. The protected group-based perturbation on input $\x$ would change \textit{only} the gender dimension of $\x$ and none of the other dimensions of $\x$.  

This kind of perturbation is relevant to checking and optimizing for fairness across groups. Let us suppose that our explanation is faithful to the underlying model. If it turns out that the explanations for the original and perturbed inputs are different when the model was not supposed to be using the protected attribute, then this is a sign that the model may be unfair. \citet{FairnessDai2021} define \textbf{counterfactual-fairness} as the ability of an explanation to appropriately reveal whether the model is unfair or not. Specifically, their mathematical formulation of sensitivity to protected group-based perturbations states that the change in explanation (according to some distance metric) must be approximately equal to the change in model output, when the input is subjected to a protected group-based perturbation:    
\begin{flalign}\label{dai_ct_fair}
&&E(f,\x) - E(f,\x') \approx f(\x) - f(\x')  && \triangleright\,\text{Equation 2.1.2 in \citeauthor{FairnessDai2021}}
\end{flalign}

In terms of optimization, \citet{FairnessDai2021} add a penalty term to LIME \citet{LIME}
to encourage counterfactual fairness: 
\begin{flalign}\label{dai_lime_fair}
&&\text{fair-LIME-loss}(E, f, \x) \defines \mathcal{L}\left(E, f, \mathcal{R} \right)+\lambda_{1}\Omega(E, \x)+\lambda_{2}\psi(f,E)  && \triangleright\,\text{Equation 2.2.2 in \citeauthor{FairnessDai2021}}
\end{flalign}
where $\mathcal{R}$ is the local neighborhood region of an input $\x$, $\lambda_{1}$ is the tuning parameter for the complexity of the explanation $\Omega$, and $\lambda_{2}$ is the tuning parameter for the fairness-preservation term
$\psi$. Here, $\psi$ will be a regularizing term which penalizes unfairness, i.e. its value decreases with increase in \textbf{counterfactual-fairness}. Other appropriate fairness metrics can be used instead of \textbf{counterfactual-fairness} as well.

\citet{Dandl2023Hyperbox} define a \textit{hyperbox} as a region of interest where we perform group perturbations of both kinds -- perturbing individual attributes and perturbing subsets of attributes at a time -- and inspect model decisions in order to explain them. It is constructed such that the model predictions for all perturbed points $\x'$ within the hyperbox are similar to $f(\x)$, and it acts as an explanation at the input $\x$. Because the hyperbox is merely a window into model decisions, model sensitivity (to group perturbations) is therefore equivalent to explanation sensitivity. The extent to which the model outputs for perturbed inputs are similar to $f(\x)$ is a measure of how sensitive the model (and hence the explanation) is to these group-based input perturbations. To measure this, \citet{Dandl2023Hyperbox} define \textbf{hyperbox-precision}. For a training set $\mathcal{X}$ such that $\x \in \mathcal{X}$, let the perturbation region covered by the hyperbox be $\mathcal{R}$ and $\lambda$ be a user-defined threshold:

\begin{flalign}\label{hyperbox_precision}
&&\text{hyperbox-precision}(E, f, \x) \defines \frac{\sum_{\x' \in \mathcal{X}} \mathds{1}\{ (\x' \in \mathcal{R}) \wedge (f(\x') - f(\x) < \lambda) \}}{\sum_{\x' \in \mathcal{X}}{\mathds{1}\{\x' \in \mathcal{R} \}}} && \triangleright\,\text{Equation 3 in \citeauthor{Dandl2023Hyperbox}}
\end{flalign}

\subsection{Sensitivity to Model Perturbations}\label{subsec:modelparameterperturb}

So far, we have focused on mathematical formulations of sensitivity that describe how the explanation changes with respect to changes in the input. However, other kinds of sensitivities may be of interest as well. For example, consider a perturbation region which comprises varying models as opposed to inputs. Even though it is no surprise that perturbing the model itself changes the explanation meaningfully (after all, its purpose is to explain the model assigned to it), it is important to be able to mathematically measure this change. This helps us understand how explanations respond to changes in models, while also helping tackle undesirable explanation manipulation via adversarial model perturbations.

\paragraph{Model Parameter Perturbations.}
\citet{Adebayo2018} measure the sensitivity of feature attribution explanations to model parameters (in the context of images). Specifically, their \textbf{model-parameter-sensitivity} formulation measures the difference between the explanation for a model with learned weights and the explanation for an untrained model with the same architecture but randomly initialized weights. 
\begin{flalign}\label{modelparamsens}
&&\text{model-parameter-sensitivity}(E, f, \x) \defines \ell(E(f,\x), E(f',\x)) && \triangleright\,\text{Section 3 in \citeauthor{Adebayo2018}}
\end{flalign}
Here, $f'$ denotes the `perturbed' model with randomly initialized
weights. Ideally, this quantity should be large (unlike previous cases), because we want the explanation to meaningfully depend on the parameters of the model it is trying to explain, i.e. it \textit{should} be sensitive to the model parameters. $\ell$ can denote any distance metric -- the authors use \textbf{structural-similarity-index}, \textbf{pearson-correlation-coefficient} and \textbf{spearman-rank-correlation}, described in Section~\ref{subsec:exp_similarity}. 

In total, \citet{Adebayo2018} present three model perturbation techniques. First, they perturb a trained model $f$ by randomizing all its learned weights, thus effectively yielding an untrained model of the same architecture. Second, they perturb $f$ by re-initializing its individual layers to random weights. Re-initializing each layer yields a corresponding $f'$. Third, they progressively re-initialize layers, starting from the last layer, moving up to the first -- adding one layer at a time to the set of randomized layers. 


Many works have reported explanation insensitivity to model parameters, which underscores the importance of this property in practice. For instance, in the context of images, if specific weights (neurons) in the model are responsible for capturing relevant concepts in the image, then the presence or absence of these weights should significantly impact a pixel-based feature attribution explanation. In general, pixel-based explanation similarity can be inspected visually or via appropriate metrics described in Section~\ref{subsec:exp_similarity}. \citet{Mahendran2016SalientDN} find that backpropagation-based feature attribution explanations, which are meant to reveal concepts captured by individual weights (neurons), in fact do \textit{not} produce drastically different visual outputs upon selecting different neurons or subsets of neurons to explain. That is, such explanations are insensitive to model parameters---at least in the context of images.




In contrast to explaining neurons via backpropagation, \citet{simonyan2014deep, Mahendran2014, Yosinski2015} pose the problem of explaining model weights (neurons) or model outputs (which simply comprises the final layer of neurons), as an optimization problem. They find an image $\x'$ such that it maximizes the activation of the chosen neuron or model output for a given input $\x$. Let $f(\x)$ be the model output, then $\x' = \argmax_{\x} f(\x) - \lambda \left\Vert \x \right\Vert ^ 2$. In this case if $f(\x)$ is truly capturing a relevant concept, then $\x'$ should contain the correct concept present in the input image $\x$. Explanations generated in this manner are indeed found to be sensitive to model parameters--the image $\x'$ differs depending on the chosen neuron or model output. This indicates that in practice, explanation sensitivity to model parameters is very dependent on the type of explanation being used and its construction.

\paragraph{Adversarial Model Parameter Perturbations.}
\citet{Heo2019AdvModelManip} manipulate model weights as part of an adversarial fine-tuning step, with the goal of retaining model robustness but generating sensitive pixel-based feature importance explanations. Let the validation set be denoted by $\mathcal{X}_{val}$, the $i^{\text{th}}$ output logit for the unperturbed model $f$ be $f(\x)_{i}$ and the $j^{\text{th}}$ output logit for the perturbed (fine-tuned) model $f'$  be $f'(\x)_{j}$, such that $j \neq i$. Ideally, we would want high similarity between explanations for the same logit, from any two models. However, from an adversarial standpoint, we obtain a sensitive explanation if the explanation for the logit $f'(\x)_{j}$ is similar to the explanation for a different logit from the unperturbed model $f(\x)_{i}$. Deploying the adversarial model $f'$ would then produce misleading explanations. Let \textbf{spearman-rank-correlation} be denoted by $\mathcal{SRC}$ (see ~\ref{subsec:exp_similarity}) and the output logit $i$ be the one we want to explain. We can quantify a loss term $\mathcal{L}$ which has lower values for higher degree of adversarial fooling, as below.

\begin{equation*}
\mathcal{L}(f, f', E, f(\x)_{i}, f'(\x)_{j}) = \mathcal{SRC}(E(f',\x, f'(\x)_{i}), E(f,\x, f(\x)_{j})) - \mathcal{SRC}(E(f,\x, f(\x)_{i}), E(f',\x, f'(\x)_{i}))
\end{equation*}

\citet{Heo2019AdvModelManip} then define \textbf{adversarial-sensitivity} as below, for a user-defined threshold range $[\lambda_1, \lambda_2]$:

\begin{equation*}
\text{adversarial-sensitivity}(\E, f, \x) \defines \frac{1}{|\mathcal{X}_{val}|} \sum_{\x \in \mathcal{X}_{val}}{\mathds{1}
\{\mathcal{L}(f, f', E, f(\x)_{i}, f'(\x)_{j}) \in [\lambda_1, \lambda_2 ]\}}
\end{equation*}
\begin{flalign}\label{foolingstability}
&& \triangleright\,\text{Equation 6 in \citeauthor{Heo2019AdvModelManip}}
\end{flalign}

From a user standpoint, we want \textbf{adversarial-sensitivity} to be as low as possible, indicating the explanation's robustness to adversarially perturbed models.

\paragraph{Training Label Perturbations.}
Analogous to \textbf{model-parameter-sensitivity}, \citet{Adebayo2018} introduce \textbf{training-label-sensitivity} as the explanation's sensitivity to a model $f'$ with the same architecture as $f$, but trained with randomly permuted labels. By measuring the difference in explanations for $f$ and $f'$, we can estimate the effect of the training labels on the explanation. The mathematical formulation for \textbf{training-label-sensitivity} is the same as Equation \ref{modelparamsens}. This quantity should also ideally be large if the explanation is meaningfully dependent on the model.

Within the context of images, various explanation types appear to have differing sensitivities to training label perturbations. For example, gradient-based feature attribution explanations such as SmoothGrad \citet{Smilkov2017} and GradCAM \citet{Selvaraju2016} have been seen to have high sensitivity \citet{Adebayo2018}, which is desirable. In contrast, other gradient-based explanations such as Guided Backpropagation \citet{springenberg2015striving}, integrated gradients \citet{AxiomaticAttribution} and DeepLIFT \citet{Shrikumar2017} give visually ambiguous explanations for randomized labels, which could be misleading to an unsuspecting user. \citet{Adebayo2018} suspect that a number of factors could be contributing to this -- the choice of model architecture acting as a prior, the input dominating the input-gradient product thus diminishing the effect of the gradient (which makes the input-label connection), some explanations acting as edge detectors rather than truly model-sensitive explanations.

\paragraph{Increasing Explanation Robustness via Model Perturbations.}
\citet{Bykov2021NoiseGrad} extend the technique introduced by SmoothGrad \citet{Smilkov2017} and show empirical evidence of an increase in explanation robustness via model parameter perturbations. The intuition behind this mechanism is that if the model's decision boundary happens to be steep, the gradients in that region are bound to be noisy. By considering the decisions of several sampled models, we get a stronger signal in the steep region. This signal yields more gradient information, which when averaged gives a robust explanation. The same intuition applies to SmoothGrad too, except the perturbed quantity here is $\x$.

\subsection{Sensitivity to Explanation Hyperparameters}\label{subsec:exp_hyperparams_sens}

Sometimes, the explanation generation process itself involves the usage of hyperparameters. This is especially true in the context of images -- for instance, feature attribution explanations such as SmoothGrad \citet{Smilkov2017} smoothen noisy gradient explanations with a Gaussian kernel and introduce two hyperparameters in the process: the number of samples $|\mathcal{R}|$ in the neighborhood $\mathcal{R}$ of input $\x$, and the standard deviation $\sigma$ of the kernel $\mathcal{N}(0, \sigma^2)$. Ideally, explanations should be insensitive to the hyperparameters used to generate them.

\citet{Bansal2020} demonstrate that feature attribution explanations are in fact highly sensitive to the choice of explanation hyperparameters. This is undesirable because it negatively impacts faithfulness, user trust and explanation reproducibility. The authors use the explanation similarity metric \textbf{structural-similarity-index} (Section ~\ref{subsec:exp_similarity}) -- a common choice in the context of images.

A common approach to computing the importance of a feature $k$ for feature-attribution explanations involves comparing the output $\hat{y}$ for the input $\x$ to the output $\hat{y}_\S$ for an input $\x_\S$ where the values of the features of $\x$ not in $\S$ are reverted to baseline values $\x_0$: $\x_\S^{(k)}=\x^{(k)}$ if  \, $k \in \S$; otherwise $\x_\S^{(k)} = \x_0^{(k)}$. This choice of baseline $\x_0$ is a hyperparameter for computing the explanation. 
Some methods opt to set it to an arbitrarily chosen value of zero, which has been used to measure faithfulness via no false positives, called \textbf{missingness} (see Equation~\ref{missingness}). \citet{sturmfels2020visualizing, Kindermans2017} demonstrate that the explanation heavily depends on the choice of this baseline value. \citet{Kindermans2017} also find that sensitivity to the choice of baseline $\x_0$ also decreases \textbf{input-variance} (Section \ref{subsec:advperturb}). That is, the explanation's sensitivity to the baseline hyperparameter $\x_0$ directly impacts its sensitivity to input perturbations. In general, the appropriate choice of baseline $\x_0$ will be task and domain dependent.

\subsection{Examples of Explanation Similarity Metrics}\label{subsec:exp_similarity}


All mathematical formulations for explanation sensitivity require some notion of distance between explanations. In general, the appropriate distance will be domain and explanation-type dependent. Below, we describe common similarity metrics to compare the explanations for images. In all the similarity metrics, the explanations are first normalized such that their weights sum to one.

The metric \textbf{structural-similarity-index}, used by \citet{Adebayo2018, Bansal2020} and introduced by \citet{Wang2004} is formulated as:
\begin{equation*}
\text{structural-similarity-index}(E_{\x}, E_{\x'}) \defines \frac{(2\mu_{E_{\x}}\mu_{E_{\x'}} + C_{1})(2\sigma_{E_{\x}E_{\x'}} + C_{2})}{(\mu_{E_{\x}}^2 + \mu_{E_{\x'}}^2 + C_{1})(\sigma_{E_{\x}}^{2} + \sigma_{E_{\x'}}^{2} + C_{2})} 
\end{equation*}
\begin{flalign}\label{ssim_sens}
&& \triangleright\,\text{Equation 13 in \citeauthor{Wang2004}}
\end{flalign}
where $\E_{\x}$ and $\E_{\x'}$ represent $\E(f,\x)$ and $\E(f,\x')$ respectively and are used for notational brevity.
Let $E_{\x_{i}}$ be the $i^{th}$ pixel in the explanation $E_{\x}$.
\citet{Wang2004} call the average over the $N$ pixels in $E_{\x}$, $\mu_{E_{\x}} = \frac{1}{N} \sum_{N} E_{\x}$, the \textbf{luminance}. They call the corresponding standard deviation $\sigma_{E_{\x}} = \frac{1}{N-1}\sum_{N} (E_{\x_{i}} - \mu_{E_{\x}})^2$ the \textbf{contrast}. $C_{1}$ and $C_{2}$ are constants which ensure a non-zero denominator.

Another commonly-used similarity metric is the \textbf{pearson-correlation-coefficient} (PCC) of the histogram of gradients (HOGs).  The HOG feature vector $\vec{H_{\x}}$ is
.  The PCC ($\rho$) is the cosine of the angle between these vectors:
\begin{flalign}\label{pearson_hog_sens}
&&\text{pearson-correlation-coefficient}(\vec{H_{\x}}, \vec{H_{\x'}}) = \rho(\vec{H_{\x}}, \vec{H_{\x'}}) = \frac{\vec{H_{\x}} \cdot \vec{H_{\x'}}}{\lVert \vec{H_{\x}} \rVert \lVert \vec{H_{\x'}} \rVert} &&
\end{flalign}

Another metric is the \textbf{mean-squared-error} (MSE) which is the absolute difference between the explanations $E_{\x}$ and $E_{\x'}$.
\begin{flalign}\label{mse_sens}
&&\text{mean-squared-error}(E_{\x}, E_{\x'}) = \frac{1}{N} \mathlarger{\sum}_
sopster
{i=1}^{N} (\lvert E_{\x_{i}} - E_{\x'_{i}} \rvert)^{2} &&
\end{flalign}

\citet{russell2012spearman} introduced \textbf{spearman-rank-correlation}, which has been used by \citet{Adebayo2018, Ghorbani2017NNFragile, Heo2019AdvModelManip, Singh2019AttributionalRobustness}. Consider a feature importance explanation (in this context, pixel-based). The \textit{rank value} of each pixel in the explanation is equal to its relative position compared to all other pixels, based on numerical value. Consider two explanations $E_{\x}$ and $E_{\x'}$ which have $K$ components each. Let the \textit{rank value difference} between corresponding components of the two explanations be $E_{\x_{k}} - E_{\x'_{k}}$ for the $k^{th}$ component. Then the mathematical formulation is: 
\begin{flalign}\label{spearman_sens}
&&\text{spearman-rank-correlation}(\E(f,\x), \E(f,\x')) = \mathcal{SRC}(\E(f,\x), \E(f,\x')) = 1 - \frac{6 \mathlarger{\sum}_{k=1}^{K} (\E_{\x_{k}} - \E_{\x'_{k}})^{2}}{K(K^{2} - 1)} &&
\end{flalign}



\citet{Ghorbani2017NNFragile, Chen2019Regularization} use \textbf{top-$k$-intersection}, which is useful in settings where the top $k$ most important features are of interest. It is defined as the size of the intersection of two sets -- $\S_\x$ and $\S_{\x'}$, which denote the top-$k$ retained features by $\E(f,\x)$ and $\E(f,\x')$ respectively.

\begin{flalign}\label{topkintersection}
&&\text{top-$k$-intersection}(\E(f,\x), \E(f,\x')) = |\S_\x \cap \S_{\x'}| &&
\end{flalign}

\citet{Kendall1938RankCorrelation} introduced \textbf{kendall-rank-correlation}, which has been used by \citet{Singh2019AttributionalRobustness, Chen2019Regularization}. 

\begin{flalign}\label{kendall_sens}
&&\text{kendall-rank-correlation}(\E(f,\x), \E(f,\x')) = \tau(\E(f,\x), \E(f,\x')) = \frac{C - D}{C + D} &&
\end{flalign}

\citet{Etmann2019Connection} measure distance between the input (image) $\x$ and the explanation $E(f,\x)$ using a \textbf{dot product}. This notion of distance is used in defining explanation sensitivity as \textbf{alignment} ~\ref{alignment}.

\citet{Singh2019AttributionalRobustness} measure similarity between an explanation $E(f,\x)$ and its input (image) $\x$ using \textbf{cosine-similarity}, which is defined as:

\begin{flalign}\label{cosine_sim_sens}
&&\text{cosine-similarity}(\E(f,\x), \x) = \cos{(\E(f,\x), \x)} = \frac{\E(f,\x) \cdot \x}{\left\Vert \E(f,\x), \x) \right\Vert \cdot \left\Vert \x \right\Vert} &&
\end{flalign}

An analogous metric that is derived from \textbf{cosine-similarity} and measures dissimilarity (higher values for dissimilar quantities) is \textbf{cosine-dissimilarity}:

\begin{flalign}\label{cosine_dissim_sens}
&&\text{cosine-dissimilarity}(\E(f,\x), \x) = 1 - \text{cosine-similarity} &&
\end{flalign}




\section{Faithfulness and Fidelity} \label{sec:faith}





In the literature, \emph{fidelity} and \emph{faithfulness} are used to describe how well an explanation captures the true underlying behavior of the model. 
An explanation that does not faithfully reflect the model's behavior maybe misleading, when the user must determine whether the model's output can be trusted.  
\paragraph{Existing Fomulations of Faithfulness in Literature}
While it is intuitive to define a faithful explanation as one that accurately mirrors the underlying model, faithfulness can be formalized in many different ways.  
For explanations that can be used to simulate the model's output, we generally define faithfulness as \emph{approximation performance}. That is, we measure how well an explanation approximates the model, that is, we compute the ``approximation loss", 
$$\sum \mathcal{L}(\hat{y}_f, \hat{y}_\E) = \sum_{\x \in \mathcal{X}} \mathcal{L}(f(\x), E(f,\x)(\x)),$$ 
 over a set of inputs of interest $\mathcal{X}$. Here, $\mathcal{L}$ denotes the loss that quantifies the difference between the model's prediction and the prediction given by the explanation. The predicted output from the explanation $\hat{y}_\E=E(f,\x)(\x)$ could come directly from a function-based explanation or from some heuristic overlaid on a feature attribution or example-based explanation. We describe these in Section~\ref{subsec:approxacc}.

For feature-attribution and example-based explanations (as well as concept bottleneck models), definitions of faithfulness are more diverse in their formalization, but overall they measure changes in the function output as one perturbs the input along dimensions identified as important by the explanation. Thus, we categorize these faithfulness metrics as the result of design choices for producing the input perturbations. 

The first choice is whether we choose to perturb a fixed set $S$ of potentially important input dimensions. Fixing a set $S$, we choose either to study one feature at a time or the entire set of features in $S$; we also decide whether we remove these feature values in our perturbations, or we remove the values of the complement set of features. 

These choices lead to three categories of faithfulness metrics. 
The first category defines faithfulness as \emph{no false negatives}, which evaluates the degree to which an explanation is able to detect truly important features/samples or contains important concepts. We describe these in Section~\ref{nfn}. The second defines faithfulness as \emph{no false positives}, which evaluates how well an explanation identifies truly insignificant features or samples as insignificant. We describe these in Section~\ref{nfp}. Finally, we can define faithfulness as checking for both \emph{no false positives and false negatives}, which evaluates how well an explanation can detect both truly important features as well as truly insignificant features. 

Finally, after choosing how we perturb the input, we must choose how to quantify the function's change to input perturbations: measuring the correlation between the function output and the explanation, or using the explanation to anticipate the absolute change in the function. For each metric, we point out how function change is quantified.



\subsection{Faithfulness for Function-Based Explanations} \label{subsec:approxacc}

Function-based explanations are functions designed to approximate the underlying model, either globally or in a region around a specific input point. Thus, when measuring the faithfulness of a function-based explanation, we can use one of the many metrics that we have for quantifying function fit.  For example, when the underlying model is a classifier, metrics like false positive and false negative rates can be used when we care about the explanation's errors in an asymmetric manner; metrics like accuracy or AUC (between the explanation and the model) capture the overall quality of the approximation in both directions. Below we highlight some function-based definitions of faithfulness.  We emphasize, however, when it comes to the development and understanding of metrics, there is nothing specific here to explanations.  At the end of the day, we are simply comparing two functions to each other.

\citet{FairnessBalagopalan2022} adopt a general definition called \textbf{loss-based-fidelity} from \citet{Craven1995} and average the quality of approximation around all inputs $\x \in \mathcal{X}$ as:
\begin{flalign}\label{balagopalan_exp_fidelity}
&&\text{loss-based-fidelity}(E, f)  \defines \frac{1}{\lvert \mathcal{X} \rvert}\sum_{\x \in \mathcal{X}} \mathcal{L}\Big( f(\x), E(f)(\x) \Big)  && \triangleright\,\text{Definition 3.1 in \citeauthor{FairnessBalagopalan2022}}
\end{flalign}
The above captures a very general notion: simply comparing the outputs of two functions over some input space.  The choice for the loss $\mathcal{L}$ can be any metric: mean squared error, accuracy, false positive rate, etc..  If the explanations are local, then we can modify it to measure the average degree of faithfulness of each local explanation $\E(f,\x)$ at each input $\x$. \citet{lundberg2017unified} (SHAP) do exactly this, defining an explanation as \textbf{locally-accurate} if the function and the explanation match in value, that is, $f(\x)-E(f,\x)(\x)=0$.


In \citet{Yeh2019OnT}, the proposed metric, \textbf{local-infidelity}, is an instantiation of Equation~\ref{balagopalan_exp_fidelity}.  Specifically, the authors quantify the change in model output (using $\ell_2$-distance) against the change in explanation when the input is perturbed: $\sum_{\x'} \ell_2(f(\x')-f(\x), \hat{y}_\E(\x')-\hat{y}_\E(\x))$. (Recall that $\hat{y}_\E$ is the approximation of $y$ given by the explanation). This measures if the explanation---as an approximation to the model---changes in the same way as the model does locally around an input $\x$. When the explanation is a linear approximation to the model at $\x$, this becomes: 
\begin{equation*}
\text{local-infidelity}(E,f,\x, p) \defines \mathbb{E}_{p(\x')}\left[ \left((\x' - \x)^T \E(f,\x) - (f(\x) - f(\x')) \right)^2 \right] 
\end{equation*}
\begin{flalign}\label{eqn:LocalInfidelity}
&& \triangleright\,\text{Definition 2.1 in \citeauthor{Yeh2019OnT}}
\end{flalign}
where the perturbations $\x'$ are drawn from some $p(\x')$ centered at the input of interest $\x$ (rather than over the whole input space). We note that the popular explanation method LIME \citep{LIME} generates explanations that are \emph{optimal} with respect to the local-infidelity metric in Equation~\ref{eqn:LocalInfidelity}. 

These input perturbations can be performed on structured inputs.  For example, \citet{Wang2020FaithDeepGraph} use the idea of local-infidelity in Equation~\ref{eqn:LocalInfidelity} to evaluate explanations for Graph Neural Networks (GNNs). In this case, the input of interest is some graph $(X,A)$ with nodes $X$ and adjacency matrix $A$.  \citet{Wang2020FaithDeepGraph} define perturbation distributions $p_X(X')$ centered at $X$ and $p_A(A')$ centered at $A$, and then define \textbf{graph-unfaithfulness} as 
\begin{equation*}
\text{graph-unfaithfulness}(E,f, X, A, p_X, p_A) \defines 
\end{equation*}
\begin{flalign}\label{generalunfaith}
&& \mathbb{E}_{p_X(X'), p_A(A')} \left[ \big( E(f,X,A) - E(f, X', A')  - ( f(X,A)-f(X', A')) \big)^2 \right] \triangleright\,\text{Definition 3.1 in \citeauthor{Wang2020FaithDeepGraph}}
\end{flalign}
The above is exactly the same as local-infidelity in Equation~\ref{eqn:LocalInfidelity} except with graphs as the inputs.


  

The general form of Equation~\ref{balagopalan_exp_fidelity} has also been adapted to discrete approximations and outputs.  For example, \citet{Lakkaraju2017} describe how to approximate a function with a two-level decision set as the explanation.  A decision set is a collection of conditions called decision rules, with each decision rule resulting in some output (e.g. the predicted class).  An input may trigger several multiple decision rules.  \citet{Lakkaraju2017} incorporate this aspect of how decision sets work in their definition of \textbf{disagreement}, which counts how often the output of the decision set explanation $E(f)$ matches the underlying function $f$ across all the rules in $E(f)$:  
\begin{equation}
\text{disagreement}(E, f, \mathcal{X}) \defines \sum_{m=1}^M \Big\lvert \{ \x \in \mathcal{X} \vert f(\x) \neq E(f)(\x) \wedge E(f)(\x) = c_{m}\}\Big\rvert  \nonumber
\end{equation}
\begin{flalign}\label{eqn:disagreement}
&& \triangleright\,\text{Table 1.1 in \citeauthor{Lakkaraju2017}}
\end{flalign}
where $M$ is the number of decision rules in the decision set and $c_m$ is the output associated with decision rule $m$.

\subsection{Faithfulness for Feature Attribution Explanations}
For function-based explanations, we could leverage the myriad metrics that exist for comparing two functions to each other.  However, feature attribution explanations do not come with a natural mapping to function approximation and thus these metrics do not apply.  

In this section, we summarize the many new metrics that have been proposed to evaluate the faithfulness of a feature attribution.  All of these metrics contain some notion: checking whether the subset of input dimensions marked as important by the explanation have a significant impact on the function's behvariour (similarly, whether dimensions marked as unimportant are actually insignificant to the function).  This check normally involves some type of perturbation, setting certain dimensions of the input to the values of some baseline input $\x_0$.  We emphasize that this choice of baseline $\x_0$ will have a significant effect on the output of the metric: choosing an appropriate baseline is an important parameter of the metric.

Below, we divide methods into whether they consider a fixed subset of features or sweep over different subsets of features.  Within each way of choosing subsets, we further describe whether the faithfulness metric focuses on checking for the significance of features marked as important by the explanation (no false positives), on checking for the insignificance of features marked as unimportant (no false negatives), or both.

\subsubsection{Faithfulness with respect to Fixed Subsets $S$}
\label{sec:faithful:fixed_S}
The first set of metrics assume that we have some fixed subset $S$ of important dimensions, usually defined by some $s$-proportion of the most important features (as defined by the top weights in the feature attribution). 

\paragraph{Necessity and No False Positives}
Faithfulness as \emph{no false positives} evaluates whether or not an explanation can identify truly significant features; we don't want to include in the explanation features that are unimportant to the function's prediction (i.e. false positives). We summarize and compare a number of metrics that differ in how they measure the impact of features on the function.

\citet{eraser} proposes a metric, called \textbf{comprehensiveness}, that checks for false positives. They do so by discarding information contained in the most important input features to see if the function's output at the perturbed input is significantly different than its output at the unperturbed one. 
More formally, let $\x_{\E /s}$ denote the input $\x$ where $s$ proportion of the most important features are set to the values of a baseline input $\x_0$:
\begin{equation}
\x_{\E /s}^{(k)}=\x_0^{(k)} \,\text{if}\, k\in \{\text{Rank}_{\E}(f, \x)_1 \dots \text{Rank}_{\E}(f, \x)_{\ceil*{sK}}\}\,\text{else}\, \x^{(k)} \,
\end{equation}
Then, comprehensiveness is defined as:
\begin{flalign}\label{compre}
&&\text{comprehensiveness}(f, \x, E, s) \defines |f(\x) -  f(\x_{\E /s})| \, && \triangleright\,\text{Equation 2 in \citeauthor{eraser}, Equation 1 in \citeauthor{ContrastExp2019}}
\end{flalign}
Here, a higher value means that fewer unimportant features are incorrectly recognized as important, which indicates higher faithfulness. 


\citet{contrastive} and \citet{ContrastExp2019} checks for the minimality of the set of features identified as important (if the set of important features is minimal then it does not include false positives). They define \textbf{pertinent-negatives} as the minimal set of features that can be perturbed in order to bring the model to a desired output.
\cite{eval_robust_hsieh} proposes a metric called \textbf{subset-robustness}, which measure the minimum perturbation needed in a feature subset to change the model classification. Formally, \textbf{subset-robustness} is defined as follows: 

\begin{equation}
\text{subset-robustness}(E, f, \x, s) \defines \min \{ \Vert \boldsymbol{r} \Vert \ \vert \ f(\x+\boldsymbol{r}) \neq f(\x), \ \boldsymbol{r}_i=0 \text{ for } i \notin \{\text{Rank}_{\E}(f, \x)_1 \dots \text{Rank}_{\E}(f, \x)_{\ceil*{sK}} \} \}
\end{equation}

Here, $f(\x)$ and $f(\x+\boldsymbol{r})$ are the classifications of the model $f$ at the original input $\x$ and the perturbed input $\x+\boldsymbol{r}$ respectively. In $\x+\boldsymbol{r}$, only the important features (i.e. top $\ceil*{sK}$ feature) are perturbed. If the top $\ceil*{sK}$ features are truly important to the model's output, then the size of the perturbation, $\|\mathbf{r}\|$, required to alter the model's classification should be small; and thus, we should get a lower \text{subset-robustness} value. 

The authors further tests the \text{subset-robustness} of different proportions of important features. They summarize this information via \textbf{AUC-robustness}, which is defined by the area under the curve of subset-robustness plotted against the subset size $\ceil*{sK}=1 \dots K$. A smaller AUC means a higher degree of \emph{no false positives}, and is equivalent to averaging the impact of discarding features (as in Equation~\ref{n-ord}). This idea of testing the importance of multiple different subsets of the top $K$ important features arises often in literature and we will expand our discussion at the end of this section.

While the above metrics measure faithfulness of general feature attribution methods, \citet{lundberg2017unified} studies the faithfulness of additive feature attribution explanations in particular. In additive feature attribution explanations, the explanation is a linear function of an simplified input $\x^{(\text{simplified})} \in \{0,1\}^M$ from the original input $\x \in \mathbb{R}^K$. Formally, $\E(f,\x)(\x) = \phi_0 + \x^{(\text{simplified})T} \E(f,\x) = \phi_0 + \sum_{i=1}^M \x^{(\text{simplified})}_i \cdot \E(f,\x)_i$, where $E(f, \x)_i$ is the coefficient and also the attribution score for the $i$-th feature in the simplified input. They present \textbf{missingness} as an axiom of faithfulness to satisfy, requiring:

\begin{flalign}\label{missingness}
&&\x^{\text{simplified}}_i = 0 \quad \Rightarrow \quad \E(f,\x)_i=0 && \triangleright\,\text{Property 2 in \citeauthor{lundberg2017unified}}
\end{flalign}

Note that, in the above, we represent a \textit{missing} feature by setting it equal to the baseline value of 0. This is because, in a linear representation, a value of $0$ has no impact to the output. Thus, we expect that missing features should receive a $0$ attribution, otherwise we would say that the explanation has produced a \emph{false positive}. This idea that an attribution explanation should give the truly unimportant features zero attribution scores is also echoed in Equation~\ref{nonsens}.

Outside of feature attribution explanations, the notion of \emph{no false positives} also works for example-based methods. Denote, by $E(f, \x)$, a set of examples selected by the explanation as most responsible for the prediction $f(\x)$; and let $E(f, \x)_i$ be the $i^{th}$ example in this set. \citet{Nguyen2020} introduce a metric called \textbf{non-representativeness} to capture explanation infidelity:

\begin{flalign}\label{nonrepres}
&&\text{non-representativeness}(E, f, \x) \defines \frac{\sum_{i} \mathcal{L}(f(\x), f(E(f, \x)_i))}{\lvert E(f, \x) \rvert} && \triangleright\,\text{Metric 2.2 in \citeauthor{Nguyen2020}}
\end{flalign}

A lower value in the above means that the model's output at the target input $\x$ is similar to it's output over the set of exemplars for $\x$. This indicates that the selected examples are representative of the model's behavior at $\x$ and, hence, the explanation does not contain false positives. 

\paragraph{Sufficiency and No False Negatives}

Faithfulness as \emph{no false negatives} evaluates the degree to which nothing important is left out in the explanation; or, alternatively, that the information identified as important by the explanation is \emph{sufficient} to understand the model. While this notion of faithfulness is most applicable to feature attribution methods---e.g. we do not want important features left out---no false negatives has also been applied to explanations derived from concept models. 

\citet{eraser} consider feature attribution methods that assign non-negative importance scores to features, with a higher value indicating a feature is more important for approximating the model. They capture the notion of faithfulness through the sufficiency of the top important features for approximating $f$. \citet{contrastive} and \citet{ContrastExp2019} introduce an identical metric called \textbf{pertinent-positives} measures the ability of a set of important features to justify a model output. 

We formalize the notion of sufficiency by focusing on a proportion of features considered to be important by the explanation. 
Let $\x_{\E_s}$ denote an perturbation of an input $\x$, where only a proportion $s \in [0,1]$, called \emph{retention proportion}, of features in $\x$ are retained, and the rest are set to some baseline value. The set of features retained in $\x_{\E_s}$ are selected from the set of features marked as most important by the explanation. 
For features indexed $k=1\dots K$, let the number of retained features be $\ceil*{sK}$; let $\text{Rank}_{\E}(f,\x)$ denote the \emph{ranking} of the $K$ features from highest to lowest and $\x_0$ is the reference (baseline) value for each feature:

\begin{equation} \label{xes_def}
    \x_{\E_s}^{(k)}=\x^{(k)} \,\text{if}\, k\in \{\text{Rank}_{\E}(f, \x)_1 \dots \text{Rank}_{\E}(f, \x)_{\ceil*{sK}}\}\,\text{else}\, \x_0^{(k)}
\end{equation} 

We define a metric called \textbf{(in)sufficiency} as a function of the proportion of retained important features $s$: 

\begin{flalign}\label{insuff}
&&\text{(in)sufficiency}(f, \x, E, s) \defines |f(\x) -  f(\x_{\E s})| && \triangleright\,\text{Equation 1 in \citeauthor{eraser}, Equation 2 in \citeauthor{ContrastExp2019}}
\end{flalign}
For this metric, a lower value means that the set of features identified as important by the explanation is sufficient for approximating the function's output at $\mathbf{x}$, and thus no important feature has been left out (i.e. falsely identified by the explanation as unimportant). 

The notion of \emph{no false negatives} can also be generalized to evaluate concept bottleneck models, which predict higher-level concepts from features, and then use these concepts to predict the target. \citet{ConceptCompYeh2019} illustrate that for classification tasks, if the difference between a model's performance using concept scores and one using the original input features is small, then the set of concepts is sufficient for understanding the model. Specifically, denote the inputs as $\x$ and the corresponding ground truth labels as $y$ in the validation set $\mathcal{V}$. Let $h$ be a mapping from the concepts to the prediction and $a_r$ be the random prediction accuracy that lower-bounds the metric score to 0. Then \textbf{concept-faithfulness} can be formalized as follows: 

\begin{flalign}\label{comple}
&&\text{concept-faithfulness}(E, f, \mathcal{V}) \defines \frac{\sup_h{\mathbb{E}_{\x,y \in \mathcal{V}} [y = h(E(f,\x))] - a_r }}{\mathbb{E}_{\x,y \in \mathcal{V}} [y = f(\x)] - a_r} && \triangleright\,\text{Definition 3.1 in \citeauthor{ConceptCompYeh2019}}
\end{flalign}

Higher values mean that fewer important concepts are not captured, which indicates more faithfulness. Note that although this metric was originally termed as `completeness', it evaluates the degree of \emph{no false negatives} and hence differs from the more fitting notion of completeness presented in Section~\ref{sec:faithfulness_absolute_change}, which measures both \emph{no false negatives} and \emph{no false positives}.

\paragraph{False Negatives and Adversarial Attacks.}
There are a number of adversarial attacks in literature that are designed to produce unfaithful explanations by increasing their false negatives. 
For example, \citet{Slack2019} demonstrates that local post-hoc explanation methods, like LIME and SHAP, are very sensitive to the distribution of input perturbations used to compute the explanation at a particular input. By generating input perturbations from specific distributions, one is able to produce LIME and SHAP explanations that have very high false negative rates. These attacks on explanation faithfulness can render them useless for tasks wherein we need to audit models for fairness \citep{Dimanov2020}.

\paragraph{Averaging as an Alternative to Choosing the Subset $S$}

When the subset $S$ is fixed, we have to make a somewhat arbitrary choice about which features are going to be considered important and which features are unimportant.  That is, in some cases, the feature attributions in the explanation may be very clear: one subset of features has high attributions, and another has attributions near zero.  However, in most cases, there may not be an obvious way to go from continuous feature attributions to a binary decision on the importance of the feature. 

Addressing this question, works like \citet{OOD} and \citet{eval_robust_hsieh} propose averaging over different values of retained proportion of features $s$ to avoid having to make an arbitrary choice. In a similar vein, \citet{Wang2020} study the faithfulness of attribution methods that assign real-valued (possibly negative) attributions to features. However, when quantifying faithfulness, they only consider features with positive attributions, $K^+$. Like Equation~\ref{insuff}, they check if retaining the most important features are sufficient for computing the underlying function $f$. The metric \textbf{sufficiency-ordering} is then formalized as follows: 
\begin{equation*}
\text{sufficiency-ordering}(E, f, \x) \defines \frac{1}{K^+ +1}\sum_{s \in \left\{ \frac{j}{K} \vert j = 0 \dots K^+ \right\}}  \min \left\{ f(\x_{E_s}), f\left(\x_{E_{\left(\frac{K^+}{K}\right)}}\right)\right\} - f(\x_0)
\end{equation*}
\begin{flalign}\label{sufforder}
&& \triangleright\,\text{Equation 4 in \citeauthor{Wang2020}}
\end{flalign}

In contrast to Equation~\ref{insuff}, higher values mean fewer important features being recognized as unimportant. Note that both formulations quantify faithfulness as the extent to which it is sufficient to use the most important features to reproduce the model's output. In Equation~\ref{insuff}, higher degrees of faithfulness correspond to smaller differences between the model output at the perturbed input, where only the most important features are retains, and its output at the baseline input. In contrast, in Equation~\ref{sufforder}, higher degrees of faithfulness correspond to larger differences between the model's output at the perturbed and baseline inputs. Equation~\ref{sufforder} also averages the impact of adding features to the baseline input, over all possible values of the retention proportion $s$ of features. Thus, faithfulness defined as sufficiency-ordering does not force us to choose a specific set of features to retain. 

In contrast to sufficiency-ordering, which retains the most important features (Equation~\ref{sufforder}), \citet{Wang2020} also propose measuring faithfulness in real-valued attributions by \emph{discarding} the most important features. The impact on the model output will then quantify the significance of those discarded features. As in Equation~\ref{sufforder}, they restrict to using the ranking of only $K^+$ features with positive attributions. The metric \textbf{necessity-ordering} is defined as:

\begin{equation*}
\text{necessity-ordering}(E, f, \x) \defines \frac{1}{K^+ +1}\sum_{s \in \{ \frac{j}{K} \vert j = 0 \dots K^+ \}} \max \{ f(\x_{E /s}) - f(\x_0), 0 \}
\end{equation*}
\begin{flalign}\label{n-ord}
&& \triangleright\,\text{Equation 3 in \citeauthor{Wang2020}}
\end{flalign}

In contrast to Equation~\ref{compre}, lower values in Equation~\ref{n-ord} indicate fewer unimportant features being mistaken for being important. 
In Equation~\ref{compre}, the impact of discarding features is indicated by a larger difference between the original output and the output at the perturbed input, where the most important features are discarded (i.e. reverted to the baseline). In Equation~\ref{n-ord}, the impact of discarding features is larger when we get a smaller difference between the baseline output and the output at the perturbed input, where the most important features are reverted to the baseline. Equation~\ref{n-ord} is also different in that it clips scores to be non-negative, and averages the impact of discarding features over all possible values of $s$, as proposed by \citet{OOD}. 

\subsubsection{Faithfulness as Relative Change as Subsets Change}
In the last part of Section~\ref{sec:faithful:fixed_S}, we noted that some metrics avoid having to choose a subset $S$ by averaging over different values of the retained proportion $s$.  A more sophisticated approach is to see how changing the retained proportion $s$ affects the faithfulness: this gives us insight into whether the \emph{ordering} of the feature importance is accurate.  (Note: when features interact to produce the output, there will still be a mismatch between feature-attribution explanations---which assign an importance to each input dimension independently---and the fact that a group of dimensions acting together were necessary to produce the output.)

In this section, we will look at metrics grounded in the idea that changing a supposedly important feature should cause a larger change than changing a supposedly unimportant feature.  We call these metrics ones that focus on \emph{relative change}: their main concern is that the feature attribution in the explanation correspond with impact on model in terms of relative ordering.  In Section~\ref{sec:faithfulness_absolute_change}, we will consider metrics that are grounded in the idea that the feature attribution score for a given feature should be equal to the actual change in the model when that feature is changed.  

\paragraph{Sufficiency and No False Positives}
In defining their metric for classifiers, \citet{Arya2019} consider setting features from the input $\x$ to some uninformative baseline, $\x_0$, working down from the feature with the highest feature attribution.  The claim is that the probability assigned to the original input's prediction should decrease as more features are set to their baseline values.  Further, replacing the most supposedly important features should have a larger decrease in probability than replacing supposedly less important features.

Specifically, \citet{Arya2019} compute the correlation between the feature importances $\E(f,\x)_i \in [0,1]$ and the classification probabilities $f(\x_{\{1 \dots K\} \backslash \{i\}})$, where for each $\x_{\{1 \dots K\} \backslash \{i\}}$, the $i^{th}$ feature value in $\x$ is replaced by a baseline value $\x_0$: 
\begin{equation*}
\text{monotonic-decrease}(E, f,\x) \defines - \underset{i \in \{1\dots K\}}{\text{corr}} \bigg( E(f,\x)_i, f(\x_{\{1 \dots K\} \backslash \{i\}}) \bigg) 
\end{equation*}
\begin{flalign}\label{mntffn}
&& \triangleright\,\text{Section 5.3 in \citeauthor{AlvarezMelis2018}, Equation 6 in \citeauthor{Arya2019}}
\end{flalign}
Higher correlation values in monotonic-decrease indicate that discarding (reverting) a feature with high importance will decrease the model's classification probability more than discarding features with low importance. Thus, monotonic-decrease tests for false positives.

\paragraph{Necessity and No False Negatives}
To check for false negatives, \citet{Arya2019, ContrastExp2019} use the complement of the approach above: instead of discarding the $i^{th}$ feature, only the $i^{th}$ feature is retained and the rest are set to the baseline.  The claim is that if a feature is truly unimportant, then retaining it should have little increase in the model's confidence towards the original class probability.  They define \textbf{monotonic-increase} as the correlation between the feature importances and the classification probabilities $f(\x_{\{i\}})$, where $\x_{\{i\}}$ denotes the input $\x$ where only the $i^{th}$ feature is retained and all other features are reset to the baseline value $\x_0$:
\begin{equation*}
\text{monotonic-increase}(E, f,\x) \defines  \underset{i \in \{1\dots K\}}{\text{corr}} \bigg( E(f, \x)_i, f(\x_{\{i\}}) \bigg)
\end{equation*}
\begin{flalign}\label{mntmnt2}
&& \triangleright\,\text{Section 4.5 from \citeauthor{Arya2019}}
\end{flalign}
Higher values indicate that a feature with higher importance influences the model's output more when added to the baseline, hence there are no false negatives.

\paragraph{Checking Both Directions}
When we consider only a fixed potentially-important subset, metrics tend to focus on identifying whether incorrect elements are in the subset (false positives) or whether elements have been left out (false negatives).  However, when we sweep over subsets, we have the chance to check more generally how changes in different subsets affect the function.  The idea of \emph{monotonicity} states that subsets with features with larger attribution weights should have more impact on the output than those with lower attribution weights.

In the following, let $\{1\dots K\}^{\ceil*{sK}}$ denote the $\ceil*{sK}$-sized subsets of the full feature set $\{1\dots K\}$. Then, \citet{Bhatt2020} define \textbf{attribution faithfulness} at proportionality $s$ as the correlation between all subsets in which $s$-proportion of features are retained and the feature importances associated with that subset.  Let $\x_{\S^c}$ denote the input $\x$ where all features in $\S^c$ are retained and features in $\S$ are reset to baseline values:
\begin{equation*}
\text{attribution-faithfulness}(f,E, \x, s) \defines \underset{\S \in \{1\dots K\}^{\ceil*{sK}}}{\text{corr}} \left( \sum_{i \in \S} E(f,\x)_i, f(\x) - f(\x_{\S^c}) \right) 
\end{equation*}
\begin{flalign}\label{monofaith}
&& \triangleright\,\text{Definition 3 in \citeauthor{Bhatt2020}}
\end{flalign}

\citet{Nguyen2020} use the same general idea in their metric \textbf{Spearman monotonicity}.  Rather than subsets of proportion $s$ that are set to the baseline, they consider only subsets of size one that are sampled randomly (that is, only one feature is perturbed at a time).  They also consider an arbitrary loss $\mathcal{L}$ between the original and perturbed function output: 
\begin{equation*}
\text{spearman-monotonicity}(E, f,\x) = \underset{i \in \{1\dots K\}}{\text{spearman-corr}} \bigg( \big\lvert E(f,\x)_i \big\rvert, \mathbb{E}_{\x_i} \big[\mathcal{L}(f(\x), f(\x_{\{1 \dots K\} \backslash \{i\}}) \big] \bigg)  \end{equation*}
\begin{flalign}\label{monotonicity}
&& \triangleright\,\text{Metric 2.3 in \citeauthor{Nguyen2020}}
\end{flalign}
Here, the expectation of the loss $\mathcal{L}$ is taken with respect to the randomly sampled $i^{th}$ feature. The explanation is a vector of real (i.e. possibly negative) attribution values, but only absolute attribution values are used in the metric to measure the feature's importance. Higher values indicate that perturbing a feature with high importance would cause more change in the model output, and this indicates higher faithfulness.  

From the two definitions above, we see the same general idea: perturbing---either via sampling, or setting to baseline---feature sets with more collective importance should have a larger effect on the function output than perturbing feature sets with less importance.  There are choices to be made about the sizes of the feature sets (proportion $s$), the type of perturbation (to some baseline or from some distribution), and how the original and perturbed output will be compared (a simple subtraction, or some other loss).  Other metrics that use this general idea to measure faithfulness include \textbf{SENN-faithfulness} \citep{AlvarezMelis2018}, which uses the forms of Equations~\ref{mntffn},~\ref{monofaith},~\ref{monotonicity} and perturbs SENN-model features by setting their associated coefficients to zero.


The above metrics use monotonicity to evaluate the fidelity of a \emph{single} explanation with respect to a particular input and model.  \citet{lundberg2017unified} use the same idea to measure the fidelity of \emph{an explanation method} across models.  Specifically, they look at two models $f_1$ and $f_2$ and their respective explanations $\E(f_1,\x)$ and $\E(f_2,\x)$ at an input $\x$ and define \textbf{monotonic-consistency} as below:
\begin{equation*}
f_1(\x)-f_1(\x_{\{1 \dots K\} \backslash \{i\}}) \geq f_2(\x)-f_2(\x_{\{1 \dots K\} \backslash \{i\}}) \quad \Rightarrow \quad E(f_1,\x) \geq E(f_2,\x)
\end{equation*}
\begin{flalign}\label{mono_consistency}
&& \triangleright\,\text{Property 3 in \citeauthor{lundberg2017unified}}
\end{flalign}
The equation above conveys the intuition that if removing a feature impacts one model's output more than another, the explanation's feature importance will be higher for that model.

Finally, various correlations are not the only way to check for both false positives and negatives. \citet{Nguyen2020} argue that an attribution explanation should give the truly unimportant features zero attribution scores. Let $\S_0$ denote the set of features assigned zero attribution, and $\S_f$ denote the set of features the model does not functionally depend on: 
\begin{equation}
\S_0(E, f, \x) \defines \{ i \in \{1 \dots K\} \vert E(f, \x)_i=0 \}
\end{equation}
\begin{equation}
\S_f(E, f, \x) \defines \{ i \in \{1 \dots K\} \vert \mathbb{E}_{\x_i} [ \mathcal{L}(f(\x), f(\x_{\{1 \dots K\} \backslash \{i\}})] = 0 \}
\end{equation}
Then, they define \textbf{non-sensitivity} as the symmetric difference between these two sets:
\begin{flalign}\label{nonsens}
&&\text{non-sensitivity}(E, f, \x) \defines \big\lvert \S_0(E, f, \x) \ \Delta \ \S_f(E, f, \x) \big\rvert && \triangleright\,\text{Metric 2.4 in \citeauthor{Nguyen2020}}
\end{flalign}
where $\Delta$ represents the symmetric difference of two sets.  This definition considers a more binary notion of fidelity, focused on whether a feature is either important or unimportant, rather than the value of the feature importance.

\subsubsection{Faithfulness as Absolute Change as Subsets Change}
\label{sec:faithfulness_absolute_change}

In the previous section, we focused on definitions of fidelity centered on the idea that changing (potentially sets of) inputs with higher feature importance should change the output more than changing inputs with less feature importance.  A collection of fidelity metrics take this one step farther: rather than expecting some proportionality or overall monotonicity between the feature importances of perturbed features and the change in the output, the metrics in this section require that the values of the feature importances associated with the perturbed features \emph{exactly} match the change in the output.  In some works, this notion of fidelity is called \emph{completeness}.  

\paragraph{Checking in Both Directions}
The metrics in this space can be seen as variations on the theme of \emph{feature-set or group-based faithfulness}: if a set of features has the same sum of attributions as another, then both sets must have the same impact on the model output. Because there is expected to be an absolute correspondence between the magnitude of the feature attribution and the magnitude of change in the output, most of these metrics can be viewed as checking in both directions: perturbing a feature or subset with a small attribution should have a small effect on the output, and perturbing a feature or subset with a large attribution should have a large effect.

\citet{Wang2020} consider two variants of feature-set faithfulness, one which validates by keeping features (sufficiency) and one which invalidates by removing features (necessity).  For both versions, they consider two subsets of features $\S_1$ and $\S_2$ such that $\S_1$ contains the top important features and $\S_2$ contains the least important features. Each subset accounts for an equal portion of the total attribution value, that is, $\sum_{i \in \S_1} E(f, \x)_i=\sum_{i \in \S_2} E(f, \x)_i = s$.  
If the proportion $s < 0.5$, then this process corresponds to \emph{keeping} features to validate; they define this case as \textbf{sufficiency-proportionality-s}: 
\begin{flalign}\label{propkN}
&&\text{sufficiency-proportionality-s}(E, f, \x, s) \defines \big\lvert  f(\x_{\S_1}) - f(\x_{\S_2}) \big\rvert && \triangleright\,\text{Definition 3 in \citeauthor{Wang2020}}
\end{flalign}
Lower values indicate that the two sets of features with identical attribution sums (i.e. importances) have a similar contribution in preserving the model output when retained, indicating higher faithfulness.  If the proportion $s > 0.5$, then this process corresponds to \emph{discarding} features; they define this case as \textbf{necessity-proportionality-s}:
\begin{flalign}\label{propkS}
&&\text{necessity-proportionality-s}(E, f, \x, 1-s) \defines \big\lvert  f(\x_{\S_1^c}) - f(\x_{\S_2^c}) \big\rvert && \triangleright\,\text{Definition 5 in \citeauthor{Wang2020}}
\end{flalign}
Here, $\x_{\S_1^c}$ denotes the input $\x$ where all features in $\S_1$ are reset to baseline values $\x_0$ i.e. $\x_{\S_1^c} = \x_0^{(k)} \, \text{if} \, k \in \S_1 \, \text{else} \, \x^{(k)}$ for $k=1 \dots K$, and a similar notation holds for $\x_{\S_2^c}$. Lower values indicate that discarding each of the sets of features with identical attribution sums results in a similar impact on the model output. This indicates higher faithfulness.

Rather than consider subsets built from the supposedly most and least important features, \citet{Shrikumar2017} consider the effect of each individual feature attribution (that is, all subsets of size 1).  They define \textbf{summation-to-delta} as saying that the sum of the feature attributions must equal the difference between the true model output and the output from a baseline input $\x_0$ can be broken down as the sum of individual feature contributions:
\begin{flalign}\label{sumtodelta}
&&\sum_{i \in \{ 1 \dots K\}} E(f,\x)_i = f(\x) - f(\x_0)&& \triangleright\,\text{Equation 1 in \citeauthor{Shrikumar2017}}
\end{flalign}
Note that for an explanation to have this notion of faithfulness, it must somehow be aware of the foil or baseline $x_0$. \citet{AxiomaticAttribution} introduce the explanation method integrated gradients and show that their method satisfies the summation-to-delta condition in Equation~\ref{sumtodelta}. 

A generalization of Equation~\ref{sumtodelta} is that this condition must hold for every subset, not just subsets of size 1. \citet{Ancona2017} call this version \textbf{sensitivity-n}: 
\begin{flalign}\label{sensN}
&&\text{ for } \S \in \{1 \dots K\}^n : \ \sum_{i \in \S} E(f,\x)_i = f(\x) - f(\x_{\S^c}) && \triangleright\,\text{Section 4 in \citeauthor{Ancona2017}}
\end{flalign}
Equation~\ref{sensN} is both computationally expensive to compute and quite strict, and only satisfied if the model is locally linear around the input and baseline \citep{Ancona2017}.  However, deviations from this condition can still be used as a measure of (in)fidelity.

Finally, this idea of matching attributions to change can also be applied to individual nodes in a network.  For example, layer-wise relevance propagation (LRP) introduced by \citep{Bach2015OnPE} decomposes the output of an image classifier into layer-wise relevances of pixels. For each neuron, a positive or negative relevance score indicates a positive or negative contribution to the output respectively.

Let $R_i^{(l)}(\x)$ represent the relevance score of the $i^{th}$ neuron in layer $l$ for an input $\x$ and $R_k^{(l+1)}(\x)$ represent the relevance score of the $k^{th}$ neuron in layer $l+1$. Let $w_{hk}^{(l \rightarrow l+1)}$ be the weight connecting each neuron $h$ in layer $l$ to the $k^{th}$ neuron in layer $l+1$, with a corresponding activation $a_{h}$. Then, the \textbf{relevance} of a neuron can be mathematically defined as the sum of the relevance scores of neurons belonging to the succeeding layer:
\begin{equation}
R_i^{(l)}(\x) \defines \sum_{k: \ i\text{ is input for neuron }k} R_{k}^{(l+1)}(\x) \frac{a_{i} w_{ik}^{(l \rightarrow l+1)}}{\sum_{h:\ h \text{ is input for neuron }k} a_{h} w_{hk}^{(l \rightarrow l+1)}}
\end{equation}
Next, layer-wise \textbf{relevance conservation} can be defined as the sum of relevance scores of neurons in each layer to be equal the model output:
\begin{flalign}\label{relevance_conserve}
&&f(\x) = ... = \sum_{i}R_{i}^{(l+1)}(\x) = \sum_{i}R_{i}^{(l)}(\x) = \dots = \sum_{i}R_{i}^{(1)}(\x) && \triangleright\,\text{Equation 2 in \citeauthor{Bach2015OnPE}}
\end{flalign}
If conservation is satisfied, $R_i^{(1)}(\x)$, the relevance score for each neuron in the first layer, can be viewed as the attribution for each feature or pixel at input $\x$, and the sum of attributions is required to equal the model output. Note that if there exists a baseline value $\x_0$ for the input $\x$ such that the model output at $\x_0$ is 0, i.e. $f(\x_0)=0$, then Equation~\ref{relevance_conserve} reduces to the same form as Equation~\ref{sumtodelta}.

\section{Complexity and Compactness} \label{sec: complexity}
In human-studies literature on interpretable machine learning, we often find that the cognitive burden of parsing explanations significantly affects the usefulness of these explanations \citep{Lage2019, Narayanan2018}. A less complex explanation will be easier for human users to understand. As a result, complexity is a commonly used measure of understandability and serves as a useful property to have in good explanations. 

Below, we describe specific formalizations of complexity.  We find that definitions of complexity tend to be specific to the explanation type: 
\begin{itemize}
\item For \emph{feature attribution explanations}, there are two common measures of complexity: the entropy of fractional contributions of features towards total importance and the minimum number of important features which retain satisfactory model performance.  We note that while feature attribution explanations are generally local to a specific input, these measures could be applied to whatever scale the explanation is for.

\item When the explanation is a \emph{continuous function-based explanation}---either the model itself or a local or global approximation to the model---one can simply consider the sparsity of the function (as measured through nonzero coefficients) or more sophisticated measures like how nonlinear it is.

\item While notions of sparsity also apply to \emph{logic-based function-based explanations}, there are specific notions of the complexity of the rules that may also be relevant to the ability of a human to understand the rule.

\end{itemize}
We expand on each of the above in the rest of this section.  While not our focus, we also note that \citet{ross2018} demonstrates that regularizing for explanation complexity during model training can ease human understanding of the learnt model, and they further show that it has the benefits of improving adversarial robustness and reducing overfitting.


\subsection{Measures of Complexity for Feature Attribution Explanations} \label{sec:localcomp}
\citet{Nguyen2020} define the \textbf{effective-complexity} of a feature-attribution explanation as the minimum number of the important features that must be retained such that the conditional expected loss over model performance does not exceed a given tolerance:
\begin{equation*}
\text{effective-complexity}(E, f, \x) \defines \argmin_{k \in \{ 1 \dots K \}} \lvert \mathcal{S}_k \rvert ,\;\; \text{ s.t. } \;\mathbb{E}_{\x_{\mathcal{S}_k^{c}}}( \mathcal{L}(f(\x), f(\x_{\mathcal{S}_k^{c}})) | \x_{\mathcal{S}_k}) < \epsilon 
\end{equation*}
\begin{flalign}\label{effcomp}
&& \triangleright\,\text{Definition 4 in \citeauthor{Nguyen2020}}
\end{flalign}
where $\epsilon$ is the loss tolerance.

The definition above presumes a feature is either included or excluded.  In contrast, \citet{Bhatt2020} take a soft view which considers the magnitudes of the attribution weights. They first define the \textbf{fractional contribution} $p_E(\x)_i$ of the $i^{th}$ feature as the relative magnitude of its attribution compared to the sum of all the attributions: 
\begin{align}\label{frac_cd_bhatt}
p_E(\x)_i \defines \frac{\lvert E(f,\x)_i \rvert}{\sum_{j=1}^K \lvert E(f,\x)_j \rvert }, \\
p_E(\x) \defines \{ p_E(\x)_1, \dots, p_E(\x)_K \}
\end{align}
Next, they define \textbf{entropy-complexity} as the entropy of the fractional contributions $p_E(\x)$:
\begin{equation*}
\text{entropy-complexity}(E, f, \x) \defines \mathbb{E}\big[- \ln \big( p_E(\x) \big) \big] = - \sum_{i=1}^K p_E(\x)_i \ln \Big( p_E(\x)_i \Big) 
\end{equation*}
\begin{flalign}\label{complexity_bhatt}
&& \triangleright\,\text{Definition 4 in \citeauthor{Bhatt2020}}
\end{flalign}
This definition is computationally convenient, but may not or many not align with how people think of complexity.



Finally, in the context of images, \citet{Nie2018, Mahendran2016SalientDN, Selvaraju2016, Samek2015} define complexity as \textbf{visual clarity}.  They argue that saliency maps based on vanilla backpropagation are visually less cleaner (more complex) than Guided Backpropagation and DeconvNet which are easily interpretable. \citet{Smilkov2017} employ \textit{explanation smoothing} to gradient-based feature attribution explanations, to obtain visually sharper features which has lower complexity than prior approaches.


\subsection{Measures of Complexity for Explanations that are Continuous Functions} \label{sec: globalcomp}
Now we consider explanations that are continuous functions---this could be a local function approximation to the decision boundary around a specific input, or the entire decision function itself.  

\paragraph{Complexity as Sparsity.}
One very common approach to measuring complexity in this case is simply sparsity, that is, the number of nonzero coefficients in the function.  

\paragraph{Complexity as Non-Linearity.}
Other notions compare the function to some very simple alternative, such as a linear function.  For example, \citet{AlvarezMelis2018} define self-explaining and inherently interpretable models which offer global explanations and are formed by progressively generalizing linear models to more complex models. They define global complexity as how far the self-explaining model is from a simple linear model.

Specifically, they introduce the notion of a self-explaining neural network (SENN).  A SENN is of the linear form $f(\x) = \sum_{i=1}^{K} E(f,\x)_{i} \mathbf{h}(\x)_{i}$, where $\mathbf{h}$ maps the $K$ original input features in $\x$ into interpretable basis concepts $\mathbf{h}(\x)$, and the interpretable coefficients $E(f,\x)$ have the descriptive capabilities of a complex model $E$. Users can understand model behavior through these coefficients. Further, this linear model can be generalized to achieve more flexibility by replacing the summation with a more general aggregation function $agg$. The linear model then becomes $f(\x) = agg(E(f,\x)_1 \mathbf{h}(\x)_1, \dots,  E(\x)_K \mathbf{h}(\x)_K)$. 

Since the coefficients serve as the explanation, the more stable the coefficients are with respect to $\x$, the closer the explanation will be to a linear model. To measure how close the coefficients at a particular input $\x$ are to a constant, \citet{AlvarezMelis2018} compute the difference between the true gradient of the function and the quantity $E(f,\x)^{T} \nabla_{\x} \mathbf{h}(\x)$. If the coefficients $E(f,\x)$ are presumed to be constant across inputs, this difference should approach zero and the model will resemble a linear model, thus minimizing complexity. The higher this difference is, the more unstable the coefficients are with respect to $\x$. Consequently, \textbf{SENN-instability} is formalized as:

\begin{flalign}\label{coefstab}
&&\text{SENN-instability}(E, f, \x) \defines \lVert \nabla_{\x} f(\x) - E(f,\x)^{T} \nabla_{\x} \mathbf{h}(\x)\rVert && \triangleright\,\text{Equation 3 in \citeauthor{AlvarezMelis2018}}
\end{flalign}

\subsection{Complexity Measures for Rule Based Explanations}
Finally, another major category of measures focuses specifically on logic or rule-based explanations. Below, we list metrics used by \citet{Lakkaraju2017} in the context of two-level decision sets; however, these concepts apply to a range of functions defined by logical formulas.

Let $E(f)$ be the decision tree learned to mimic the function $f$.  In \citet{Lakkaraju2017}, they are two-level decision sets with $M$ if-then rules, where the $i^{th}$ rule is a triple of the form $(r_{i}^{(1)}, r_{i}^{(2)}, c_i)$. Here, $r_{i}^{(1)}$ and $r_{i}^{(2)}$ are two nested if-then conditions, each comprising multiple predicates (e.g. $\text{age}\geq 50 \text{ and female}=\text{yes}$), and $c_i$ is the assigned class label if both conditions are met. The following metrics can now be defined as proxies for complexity: 
\begin{flalign}\label{lakkaraju_beta_inter_1}
&&\text{size}(E(f)) = \big\lvert E(f) \big\rvert, && \triangleright\,\text{Metric 4, Table 1 in \citeauthor{Lakkaraju2017}}
\end{flalign}

\begin{flalign}\label{lakkaraju_beta_inter_2}
&&\text{max-width}(E(f))=\max _{e \in \bigcup_{i=1}^{M}\left(r_{i}^{(1)} \cup r_{i}^{(2)}\right)} \text{width}(e), && \triangleright\,\text{Metric 5, Table 1 in \citeauthor{Lakkaraju2017}}
\end{flalign}

\begin{flalign}\label{lakkaraju_beta_inter_3}
&&\text{num-preds}(E(f))=\sum_{i=1}^{M} \bigg( \text{width }\left(r_i^{(1)}\right)+\text{width }\left(r_i^{(2)}\right) \bigg), && \triangleright\,\text{Metric 6, Table 1 in \citeauthor{Lakkaraju2017}}
\end{flalign}

\begin{flalign}\label{lakkaraju_beta_inter_4}
&&\text{num-dsets}(E(f))=\bigg\lvert \bigcup_{i=1}^{M} r_{i}^{(1)} \bigg\rvert, && \triangleright\,\text{Metric 7, Table 1 in \citeauthor{Lakkaraju2017}}
\end{flalign}

\begin{flalign}\label{lakkaraju_beta_inter_5}
&&\text{feature-overlap}(E(f))=\sum_{r_i^{(1)} \in \bigcup_{i=1}^{M} r_{i}^{(1)}} \sum_{i=1}^{M} \text { feature-overlap }\left(r_i^{(1)}, r_i^{(2)}\right) && \triangleright\,\text{Metric 8, Table 1 in \citeauthor{Lakkaraju2017}}
\end{flalign}

Here $\text{width}()$ counts the number of predicates, where a predicate refers to the feature, operator or value in a condition (e.g. ``age'', ``$\geq$'' and ``$18$'' in ``$\text{age} \geq 18$''). The number of rules in the decision set is given by its cardinality and referred to as \textbf{size}. The maximum number of unique if-then conditions ($r_{i}^{(1)}$ and $r_{i}^{(2)}$) across all rules is defined as \textbf{max-width}. The total number of predicates in all if-then conditions, inclusive of repetitions, is \textbf{num-preds}. The quantity \textbf{num-dsets} counts the number of unique outer if-then conditions. Lastly, \textbf{feature-overlap} is the number of common features in the feature space, that show up both in the inner and outer conditions of a nested if-then pair $r_{i}^{(1)}$ and $r_{i}^{(2)}$. The sum of these common features within each nested if-then condition pair, can be minimized, to minimize complexity. 

\citet{Narayanan2018} quantify complexity in decision sets in terms of analogous parameters. They use \textbf{size} (Equation~\ref{lakkaraju_beta_inter_1}) to count the number of lines in the decision set, \textbf{max-width} (Equation~\ref{lakkaraju_beta_inter_2}) to count the maximum number of terms per explanation and \textbf{feature-overlap} to count the number of repeated terms. They also define \textbf{cognitive-chunks} as the set of unique concepts used by the explanation. A decrease in any of these quantities can be used to ensure minimal complexity. It is possible to generalize all metrics discussed so far to other rule-based explanations.

\section{Homogeneity} \label{sec:homogeneity}
Homogeneity measures the sensitivity of the \emph{properties} of an explanation to input perturbations. For example, homogeneity can quantify changes in an explanation's robustness as we evaluate it on inputs from different regions of the input space.

Although the definition of homogeneity can be instantiated for any explanation property, we note that current mathematical formulations of homogeneity are focused on measuring changes in faithfulness, in the context of assessing model fairness. Many downstream tasks in fairness are concerned with preserving subgroup fairness within a population. Sensitive demographic attributes in the dataset define the subgroups within a population and a model is said to be fair if it performs identically across all subgroups, if all else is kept equal. In order to assess model fairness, an explanation has to be homogeneous -- equally faithful -- across all subgroups.

In the following, we discuss homogeneity metrics that measure changes in explanation faithfulness, but we emphasize that homogeneity can be instantiated for \emph{any} set of explanation properties.

\paragraph{Homogeneity as Group Faithfulness Difference} 
For simplicity, we consider examples where the group perturbation is protected (i.e. a single protected attribute is perturbed), but all mathematical formulations are applicable to the more general case as well.

\citet{FairnessBalagopalan2022} measure explanation homogeneity via differences in faithfulness across different groups. They borrow the definition of explanation faithfulness (Section~\ref{sec:faith} Equation~\ref{balagopalan_exp_fidelity}) from \citet{Craven1995}. Let $\mathcal{X}_{g}$ be the set of input samples belonging to a population group $g$ (e.g. females) with a sensitive attribute (e.g. sex), and $\mathcal{G}$ be the set of all population groups (e.g. \{females, males, others\}). Let the loss metric $\mathcal{L}$ be defined as AUROC, accuracy or mean error. Then \textbf{faithfulness-loss-per-group}, or \textbf{f-l-per-g} for a given group can be measured as:

\begin{equation*}
\text{faithfulness-loss-per-group}(E, f, g) = \text{f-l-per-g}(E, f, g) \defines \frac{1}{\lvert \mathcal{X}_{g} \rvert} \sum_{\x \in \mathcal{X}_{g}} \mathcal{L}(f(\x), E(f,\x))  
\end{equation*}
\begin{flalign}\label{group_exp_fidelity}
&& \triangleright\,\text{Definition 3.1 in  \citeauthor{FairnessBalagopalan2022}}
\end{flalign}

The explanation will faithfully mimic the model and yield a low $\mathcal{L}(f(\x), E(f,\x))$ only if it is sufficiently robust to population group perturbations. That is, lower values of \textbf{faithfulness-loss-per-group} indicate higher homogeneity. 

One way to compare \textbf{faithfulness-loss-per-group} across groups is to quantify the maximum degree to which an explanation's faithfulness reduces for a population group compared to the average explanation faithfulness across all groups. The maximum gap between a group's explanation faithfulness and the average faithfulness is defined as \textbf{max-faithfulness-gap}. For all groups $g \in \mathcal{G}$, we can say: 

\begin{equation*}
\text{max-faithfulness-gap}(E, f, \mathcal{G})  \defines \max _{g \in \mathcal{G}} \bigg( \text{f-l-per-g}(E, f, \mathcal{G}) - \text{f-l-per-g}(E, f, g) \bigg)
\end{equation*}
\begin{flalign}\label{balagopalan_maxfidgap}
&& \triangleright\,\text{Definition 3.3 in \citeauthor{FairnessBalagopalan2022}}
\end{flalign}

Another way to estimate homogeneity is to measure the average of the sum of pairwise differences in faithfulness for each pair of population groups $g_i, g_j \in \mathcal{G}$. This serves as a proxy for how much an explanation's faithfulness varies across groups. The smaller this value, the better. The mean faithfulness gap across groups, or \textbf{mean-faithfulness-gap} can then be defined as:

\begin{equation*}
\text{mean-faithfulness-gap}(E, f, \mathcal{G})  \defines \frac{2}{\lvert \mathcal{G} \rvert (\lvert \mathcal{G} \rvert-1)} \sum_{g_i \in \mathcal{G}} \sum_{g_j \in \mathcal{G}, j>i} \Big\lvert \text{f-l-per-g}(E, f, g_i) - \text{f-l-per-g}(E, f, g_j) \Big\rvert 
\end{equation*}
\begin{flalign}\label{balagopalan_meanfidgap_subgrp}
&&\triangleright\,\text{Definition 3.4 in \citeauthor{FairnessBalagopalan2022}}
\end{flalign}

While this work assumes that explanations with smaller faithfulness gaps are more desirable, they point to evidence suggesting that equal group performance can worsen collective welfare \citep{Corbett-Davies2018Fairness,Hu2020Fairness,Zhang2022Fairness}. An interesting direction would be to define the metrics above for different formulations faithfulness, discussed in Section~{\ref{sec:faith}}.


\section{Can We Have Them All? Tensions and Trade-offs between Explanation Properties}
A natural question is whether or not we can choose explanations that perform well across multiple properties. Unfortunately, literature has shown that no single explanation will have all properties. Thus, understanding these relationships and trade-offs can enable a user to prioritize properties of interest for their task.
In this section, we use our previous synthesis of explanation properties to identify tensions between major categories of properties. 



\subsection{Potential Tension: Faithfulness and Complexity}
\label{subsec:faithfulness-compl}
In many situations, it is possible to build an inherently interpretable model for one's task (e.g. linear regression). In this case, the explanation is both faithful (the model is its own explanation) and non-complex (otherwise it would not be inherently interpretable). However, if the explanation is not the entire model---presumably, because the model is too complex to understand---then we have a tension between faithfulness and complexity. As we make the explanation more faithful, we come closer to replicating the model perfectly, but at the cost of reducing human understandability. Others have described this tension in the context of feature attribution methods \citep{Bhatt2020}, surrogate explanations like LIME \citep{LIME} and rule-based explanations \citep{Lakkaraju2017}. 

As a specific example, \citet{LIME} loosely refer to the regularizing complexity term as the local surrogate model's complexity, such as the depth of a decision tree or the number of non-zero coefficients in a sparse linear model. 
Let the local infidelity measure be $\mathcal{L}\left(E, f, \x, \pi_\x\right)$ (Equation~\ref{eqn:LocalInfidelity} in Section~\ref{sec:faith}) and $\pi_\x$ denote a probability distribution that assigns higher weights to input points closer to $\x$, based on some distance metric. \textbf{LIME-loss} can then be formalized as:
\begin{flalign}\label{limeloss}
&&\text{LIME-loss}(E, f, \x) \defines \mathcal{L}\left(E, f, \x, \pi_\x\right)+ \lambda \cdot \Omega(E, f, \x) && \triangleright\,\text{Equation 1 in \citeauthor{LIME}}
\end{flalign}
where $\Omega(E, f, \x)$ is the sparsity penalty.

There exist several ways to mitigate this tension between faithfulness and complexity in situations where an inherently interpretable model cannot be used. The premise for local explanations like LIME is that by explaining a single prediction at a time, the explanation can be both faithful to the model locally as well as sufficiently simple to be understandable. Several works 
introduce complexity regularizers during training to help find local optima corresponding to models whose explanations are both faithful and complex \citep{LIME, ross2018, AlvarezMelis2018}. Finally, one can attempt to engage the user in a more complex, faithful explanation. For example, \citet{Bucinca2021} use cognitive forcing functions to encourage users to stay engaged with more complex information, even if doing so requires more cognitive labor.  

Some metrics make practical use of this tension between faithfulness and complexity to measurably ensure one property by applying bounds on the other. For example, in feature attributions, \textbf{effective complexity} (Equation~\ref{effcomp}) computes the minimum complexity needed to maintain an acceptable degree of faithfulness \citep{Nguyen2020}. That being said, while lower complexity allows for better understandability, it isn't always better since faithfulness can suffer when complexity drops below a threshold. This metric, serving as a toggle between faithfulness and complexity, mitigates this issue by lower bounding faithfulness. In a similar vein, \citet{Gilpin2018} underline the importance of this tradeoff by proposing that explanations should not be evaluated on a single point of this tradeoff but rather along the curve from maximum faithfulness to minimum complexity.

In the context of images, this tension has been documented by \citet{Nie2018, Mahendran2016SalientDN, Selvaraju2016, Samek2015, Bansal2020}. Explanation types like saliency maps have been found to be complex and more faithful, compared to Guided Backpropagation and DeconvNet which are minimally complex but unfaithful to the underlying model. \citet{Bansal2020} demonstrate the reduction in faithfulness for gradient-based feature attributions, when explanation smoothing (denoising) is performed.

\subsection{Potential Tension: Faithfulness and Sensitivity}
\label{subsec:faithfulness-sens}

A good explanation should ideally be sufficiently faithful to the model while being minimally sensitive to input, model parameter, label or hyperparameter perturbations. There sometimes can be a trade-off between these two properties: the degree of faithfulness can drop when we attempt to obtain a less sensitive explanation, and in the extreme case minimizing sensitivity naively would yield a constant and trivial explanation. 

For example, \citet{AlvarezMelis2018} opt for self-explaining models, where the stability of the model coefficients with respect to input perturbations is a proxy for sensitivity (\textbf{SENN-instability}, Equation~\ref{coefstab}). Note that because the model is its own explanation, model faithfulness and sensitivity are equivalent to explanation faithfulness and sensitivity respectively. In the model training objective, using \textbf{SENN-instability} as a regularization term and trading off faithfulness helps control for sensitivity of model coefficients to input perturbations. We note that despite being formally introduced as a measure for complexity, this metric measures sensitivity towards input perturbations and can be traded off with faithfulness. 


Measuring sensitivity for self-explaining models is equivalent to measuring the sensitivity of its coefficients. However, for cases where the explanation $\E(f)$ is separate from the model $f$, the explanation is often constructed as a function of the sensitivity of $f$. Such explanation functions, obtained from the model's sensitivity, have been shown to be even more sensitive to the input, i.e. unstable, than the model itself \citep{Yeh2019OnT, Lee2019, Ghorbani2017NNFragile}. In other words, the gradients of the model $f$ tend to be unstable and explanation sensitivity is often higher than (i.e. lower bounded by) model sensitivity. In such cases, a standard technique for reducing the sensitivity of an explanation in a region of interest, is to modify the explanation to be a kernel-based average of all explanations in that region. Intuitively, this corresponds to reducing the effect of outlier explanations in a region of perturbation, thus lowering explanation sensitivity. This technique, known as \textit{explanation smoothing} or \textit{denoising}, has been employed by \citet{Smilkov2017, Selvaraju2016, Shrikumar2017, springenberg2015striving}. 
However, averaging explanations over a region to reduce sensitivity can drastically reduce faithfulness \citet{Bansal2020}, which again alludes to the tension between these two properties. \citet{Yeh2019OnT} mitigate this tension by showing that smoothing an explanation under some assumptions leads to a guaranteed increase in faithfulness. In other words, robustness and faithfulness need not always have an inverse relationship. 

Another approach that has been used to address this issue of unstable gradient-based explanations is \textit{adversarial retraining}, employed by \citet{ross2018, Lee2019, Yeh2019OnT}. This technique attempts to train a model such that the learned function $f$ has smooth gradients. Smooth gradients will naturally allow for minimally sensitive gradient-based explanations. In addition to being minimally sensitive, we desire faithful explanations. How faithfully a gradient captures the underlying model $f$ depends on the curvature of $f$, i.e. a gradient-based explanation will be completely faithful to a model $f$ only when it has no curvature. A proxy for better faithfulness in this case becomes a lower Hessian upper bound which \citet{Yeh2019OnT} optimize for through training. Thus, adversarial retraining can help mitigate the tension between faithfulness and sensitivity by improving both. 

\citet{Bansal2020} show that hyperparameter perturbations to feature attribution explanations by \citet{LIME, Fong2017, zeiler2013visualizing} led to a decrease in faithfulness, measured via \textbf{top-n-localization-error} (Equation~\ref{topnlocalsens}), \textbf{deletion-curve} and \textbf{insertion-curve}.

\subsection{Potential Tension: Sensitivity and Complexity}
In the context of backpropagation-based feature attribution explanations for images, \citet{Nie2018, Mahendran2016SalientDN, Selvaraju2016, Samek2015} observe that there might be a potential tension between explanation sensitivity to training label perturbations and the visual clarity of the explanation (aka complexity). Saliency maps have been found to be very sensitive to labels but complex to understand, while methods like Guided Backpropagation and DeconvNet have been found to be minimally complex, yet insensitive to training labels. Sensitivity to labels is desirable and increases faithfulness of the explanation, which supports our findings on the trade-off between faithfulness and complexity (\ref{subsec:faithfulness-compl}).

As in previous cases, one can manage this trade-off during optimization.  For example, \citet{AlvarezMelis2018} incorporate \textbf{SENN-instability} (Equation~\ref{coefstab}) into the loss function above. More precisely, this metric is used as a regularizer in the model-optimizing objective function with the classification loss $\mathcal{L}\left(E, f, \x\right)$, where the hyperparameter $\lambda$ controls the trade-off of model performance against stability (a notion of complexity):

\begin{equation*} \label{sennloss}
\text{SENN-loss}(E, f, \x) \defines \mathcal{L}\left(E, f, \x\right) + \lambda \cdot \text{SENN-instability}(E, f, \x) 
\end{equation*}
\begin{flalign}\label{selfexpmodloss}
&& \triangleright\,\text{Section 3 in \citeauthor{AlvarezMelis2018}}
\end{flalign}

Note that in many ways, this metric is an extension of the local complexity defined by LIME \citep{LIME} for linear explanations, to global explanations.

\subsection{Potential Tension: Faithfulness and Homogeneity}
Homogeneity refers to the explanation's ability to preserve faithfulness across subgroups in the input population. This can also be interpreted as having low explanation sensitivity to subgroup perturbations. Practical applications for homogeneity are typically concerned with fairness amongst these subgroups. However, as shown by \citet{FairnessBalagopalan2022}, there can be a trade-off between homogeneity (which they refer to as fidelity gap) and the overall faithfulness of an explanation. In other words, balancing faithfulness across subgroups might decrease the overall explanation faithfulness, and optimizing for overall explanation faithfulness might result in very different degrees of subgroup faithfulness, thus leading to lower homogeneity (fairness). 
In such cases, \citet{FairnessBalagopalan2022} propose that it may be appropriate to maximize the faithfulness of the worst-case subgroup.

\section{Discussion and Conclusion} 
\label{sec:discussion}
In this paper, we have detailed the myriad mathematical formulations of four computational properties of explanations.  We focused on these properties as they are the most commonly found in the interpretability literature, and as such, also have high variation in their definitions.  However, there are other properties that future works could continue to organize.

\paragraph{Other Computational Properties}
There are more computational properties beyond the four we have discussed.  Providing a detailed explanation of how the model produces its output may inadvertently leak information about the data points used to train the model. Exemplar-based explanations are also, by their nature, not private.  The privacy literature already has many established privacy metrics, and an understanding of what each of those metrics do and do not measure.  Uncertainty 
involves exposing how sure the model is of its output; as with privacy, uncertainty quantification is an established area with established sets of measures. Translucence involves an explanation exposing the limitations of what it can and cannot be used for.  These properties either have established sets of formalizations, or are not commonly used.

\paragraph{Human-Centric Properties}
While we focus on computational properties in this survey, we note that there are another set of human-centric properties that may serve as valuable intermediate quantities between the explanation and the ultimate performance on a task by a human.  These include measures of semantic alignment between the user's mental model and the presentation in the explanation (sometimes also referred to as understandability), ways in which the explanation presentation is designed to encourage appropriate reliance, more generally ways in which the presentation of the explanation encourages engagement and learning, and how much cognitive labor is required to follow the explanation.  

To some extent, these human-centered properties do have aspects that can be made computational.  For example, \citet{fok2023search} define verifiability as the explanation's ability to help a user verify the correctness of a model's output. With the aid of user-studies, \citet{Narayanan2018} demonstrate that in rule based explanations, an increase in complexity leads to a reduction in verifiability, indicating a potential tension between them.  Thus, under some conditions, managing the computational property of complexity may assist in improving the human-centric property of verifiability, which may in turn be important for a downstream task such as ensuring that a model's decisions were made in a reasonable way.  Similarly, \cite{Bansal2020} show that high explanation sensitivity hampers its \textit{reproducibility}, what is, whether a user can simulate a model's outcome. Explanation robustness with respect to perturbations is essential for the practitioner to be able to reproduce the explanation output, which in turn impacts user trust. 

That said, unlike with computational properties, human-centric properties can only be truly measured in user studies. As with the many mathematical formalizations of computational properties, these human-centric properties have many ways in which they have been instrumented during user studies.  Organizing and standardizing them is an important human-factors (rather than computational) direction for future work.


\paragraph{Non-Properties}
Finally, we call out two properties that are sometimes noted in the literature but we argue are not really properties in the sense of being useful intermediate measures of explanation that may proxy for downstream performance on certain tasks: plausibility and convincingness.  The lack of plausibility in an explanation indicates one of two things: an error in the explanation producing method or an error in the model itself. The former results from low explanation faithfulness, while the latter indicates a problem with the model, not the explanation.  We argue that the key property for the \emph{explanation} is faithfulness, and if a faithful explanation exposes some part of the model as implausible, then the explanation has succeeded in assisting the user identify issues with the model.  Similarly, the goal of the explanation is to provide information about the model, not to convince or manipulate the user towards a particular decision.  Persuation is a fundamentally different task than model understanding.

\paragraph{Conclusion}
We have described many of the very large number of mathematical formalizations for four popular computational properties of explanations: sensitivity, faithfulness, complexity, and group-based differences. We outlined the similarities and differences between various mathematical definitions of the same property and map them to use cases. To our knowledge, is the first to address the current lack of organization in this space. We also discussed the trade-offs and relationships between these properties. 

Through this endeavor, we hope that users of explanation methods can begin to appreciate the breadth of properties that have been defined and can start to identify which specific mathematical formulation of a property may be applicable for their use case. This will aid easier usage of explanations and evaluation metrics in practical applications. We also hope that this work will help machine learning researchers be more thoughtful and intentional about what versions of properties they report when comparing a new method to existing ones.

\section*{Acknowledgments}
We acknowledge support from NSF IIS-1750358 and Institute for Applied and Computational Science at Harvard. We thank Isaac Lage and Siddharth Swaroop for their valuable feedback and suggestions.

\bibliography{references}  

\appendix

\section{Image Data-Specific Definitions of Faithfulness}

For images, one way of locating important pixels is to identify pixels that define objects. This is known as \textit{object localization} in computer vision. For this task, \textbf{top-n-localization-error} is a metric for measuring mistakes made by models that perform object identification. This metric considers the top \textit{n} object classification labels returned by the model for a given image, and counts the number of times the model does not identify \textit{any} of the objects in the image. 
Specific instantiations of this metric that have appeared in literature include top-1-localization-error and top-5-localization-error \citep{Zhou2015, Russakovsky2014}. 

Given $N$ number of images, and $M$ object labels per image predicted by the model, let $\mathcal{L}_{nm}$ be the classification loss (1 for a correct prediction and 0 otherwise) of the $n^{th}$ image's $m^{th}$ label. We can then formalize localization error as:

\begin{flalign}\label{topnlocalsens}
&&\text{top-n-localization-error}(f, \x) \defines \frac{1}{N} \sum_{i=1}^{N} \min_{j}{\mathcal{L}_{ij}} \text{ for } j \in \{1,2,...n\} && \triangleright\,\text{Equation 1 in \citeauthor{Russakovsky2014}}
\end{flalign}

As a derivative of localization-error, \citet{dabkowski2017real} introduce \textbf{minimum-saliency-region} as the minimum set of pixels (tighest crop) containing the entire salient region required for a correct classification. For this use case, let the retention proportion be $s$, computed as the fraction of retained image area. Let $\tilde{s}$ be defined as $\max{(s, 0.05)}$ to bound the minimum retention and $p$ be the probability of the correct class returned by the model when fed the cropped image. Formally, we can say: 

\begin{flalign}\label{minsaliency}
&&\text{minimum-saliency-region}(E, f, \x) \defines \log{(\tilde{s})} - \log{(p)} && \triangleright\,\text{Equation 3 in \citeauthor{dabkowski2017real}}
\end{flalign}

Since we want the retention proportion $s$ to be minimum (the tighest possible crop) and the probability $p$ to be maximum, this quantity must ideally be as small as possible. This is the difference between \textbf{top-n-localization-error} and \textbf{minimum-saliency-region}, where the latter captures only the minimum amount of relevant information present in an image, by cropping it. In contrast, \textbf{top-n-localization-error} will account for the entire object, even if many pixels are not required for the classification. A generalization to minimum-saliency-region by is \textbf{mass-center-ablation} by \citet{sturmfels2020visualizing} which computes the center of mass of the saliency map and ablates a boxed region around it. Unlike other ablations, the mass-center-ablation approach can also handle the fact that pixels are correlated by the underlying objects.

Variants of this introduced by \citet{Petsiuk2018} are \textbf{deletion-curve} and \textbf{insertion-curve}. The former captures the decrease in probability as important pixels are removed -- for a good explanation, we desire a sharp drop and low area under the curve. The metric \textbf{insertion-curve}, is complementary to deletion, and measures the increase in probability as important pixels are included -- for a good explanation, we desire maximum area under the curve. The variant by \citet{Fong2017} called \textbf{deletion-mask}, finds the smallest deletion to an image that changes the model output. 

\citet{sturmfels2020visualizing} refer to the metric deletion-curve as \textbf{top-k-ablation}. Since ablation could be introducing a distribution shift \citep{Kindermans2017}, \citet{Hooker2018} propose a variant called \textbf{remove-and-retrain}, which computationally expensive, ensures identical training and test distributions. For each image in the training and test set, this approach replaces the most important pixels with a fixed baseline value, retrains a new model on the modified training dataset and evaluates on the modified test dataset. If the attribution provided high quality importance scores, the retrained model would suffer a sharp decline in accuracy.







\section{Making Explanations Robust by Making Models Robust}
\label{app:robust_exp_via_model}

An adversarial perturbation could also result in inaccurate predictions from a sensitive model, and in turn, a sensitive explanation. However, it is intuitive to consider that an explanation, if faithful to the underlying model's behavior, would exhibit sensitivity if the model were sensitive. This inevitable trade-off between explanation faithfulness and sensitivity has been documented \citet{Tan2023RobustForFree}, making it the less interesting case. Note that when a model is sensitive, it is of little value to explain it directly. A natural strategy is to first make it robust and then ensure explanation robustness, which effectively devolves into the previously discussed case of ensuring explanation sensitivity when a model is robust. In fact, a robust model has been shown to have more robust explanations than its non-robust counterpart \citet{Tan2023RobustForFree}. We discuss this case separately, to emphasize the direct connection between model and explanation robustness as well as the extra step of making the model robust. 

A common technique for making a sensitive model more robust is \textit{adversarial retraining} where the training data includes hand-crafted adversarial examples \citet{Madry2019Resistant}. \citet{Tsipras2019RobustnessOdds} discover that such retrained robust models empirically generate more robust explanations, in contrast to regular models which require explanation smoothing for the same outcome. Another approach is \textit{model smoothing}, i.e. make the function smoother and obtaining less noisy gradients (explanations) as a result. 

\citet{ross2018} perform model smoothing by adding a regularization term penalizing high gradients of model loss $\mathcal{L}(y, f(\x'))$ with respect to the input. One can think of it as a way of ensuring that the divergence between the predictions and labels does not fluctuate rapidly for input perturbations. This has been observed to empirically generate more robust and interpretable explanations.

\begin{flalign}\label{gradient_regularizing_loss}
&&\text{gradient-regularizing-loss}(f, \x') \defines \mathcal{L}(y, f(\x')) + \lambda \cdot \left\Vert \nabla_{x} \mathcal{L}(y, f(\x')) \right\Vert ^ {2} && \triangleright\,\text{Equation 6 in \citeauthor{ross2018}}
\end{flalign}

Instead of penalizing high gradients of model loss as in \textbf{gradient-regularizing-loss}, \citet{Wang2020smoothedgeometry} instead penalize the largest change of the loss function's gradient in any direction in the model parameter space. More concretely, let the model training loss be $\mathcal{L}$, $H$ be the Hessian matrix of $\mathcal{L}$ and $\lambda$ be a penalty term which is scaled comparably with $\mathcal{L}$. Let $\xi_\text{max}$ be the largest eigenvalue of $H$, which tells us how quickly the gradient of the loss function changes with respect to changes in model parameters. We then have \textbf{hessian-model-smoothing}:

\begin{flalign}\label{hessian_model_smoothing}
&&\text{hessian-model-smoothing}(f, \x') \defines \mathcal{L}(y, f(\x')) + \lambda \xi_{\text{max}} && \triangleright\,\text{Equation 2 in \citeauthor{Wang2020smoothedgeometry}}
\end{flalign}

A major downside of retraining a model to be adversarially robust or smoother is that it costs time. It is also unclear if a retrained model is similar or drastically different from the original model \citet{Tan2023RobustForFree}. \citet{Smilkov2017, Wang2020smoothedgeometry} adopt \textbf{probabilistic-model-smoothing}, where the idea is to smoothen the gradient of a model by taking its average within a neighborhood of the point of interest. 

Given a probability distribution $p$ (for instance, \citet{Wang2020smoothedgeometry} prescribe a uniformly distributed sphere of radius $r$, centered at $\x$: $p=U(\{\x'|\left\Vert \x'-\x \right\Vert \leq r\})$), we have:

\begin{flalign}\label{probabilistic_model_smoothing_uniform}
&&\text{probabilistic-model-smoothing}(f, \x') \defines \nabla_x \mathbb{E}_p f(\x') = \mathbb{E}_p \nabla_x f(\x') && \triangleright\,\text{Definition 8 in \citeauthor{Wang2020smoothedgeometry}}
\end{flalign}

If the averaging is done by convolving (denoted by the $*$ operator) the model $f$ with an isotropic Gaussian $p$ ~ \( \sim \mathcal{N}(0, \sigma^2 I) \), then we get SmoothGrad \citet{Smilkov2017}, which is simply a variant of \textbf{probabilistic-model-smoothing} as shown by \citet{Wang2020smoothedgeometry}:

\begin{equation*}
\text{probabilistic-model-smoothing}(f, \x') \defines \mathbb{E}_p \nabla_x f(\x') = \nabla_x [(f * p)(\x')]
\end{equation*}
\begin{flalign}\label{probabilistic_model_smoothing_gaussian}
&& \triangleright\,\text{Sec 2.2 in \citeauthor{Smilkov2017}, Proposition 1 in \citeauthor{Wang2020smoothedgeometry}}
\end{flalign}

All findings in literature indicate that explanation robustness is directly tied to and influenced by model robustness. This motivates the choice of defining explanation robustness similarly to \textit{model robustness}, measured as the distance between the unperturbed input $\x$ from its nearest perturbed input $\x'$, such that the model classification changes. \citet{Etmann2019Connection} define \textbf{alignment} as proportional to the distance between the unperturbed input $\x$ and the explanation $E(f,\x)$ where the explanation is the gradient of the model: $E(f,x) = \nabla_f(x)$. Intuitively, it measures how similar the input image is to its pixel-based explanation. The smaller this distance, the more robust, i.e. \textit{aligned}, the explanation is. For linear binary classifiers, \textbf{alignment} trivially increases as model robustness increases. 

\begin{flalign}\label{alignment}
&&\text{alignment}(\E, f, \x) \defines \frac{|\x \cdot \E(f,\x)|}{\left\Vert \E(f,\x) \right\Vert} && \triangleright\,\text{Equation 3 in \citeauthor{Etmann2019Connection}}
\end{flalign}

\paragraph{Jointly Optimizing for High Model Accuracy and Explanation Robustness.}


In the context of adversarial robustness, most works retrain the model to make it more robust. In contrast, \citet{Singh2019AttributionalRobustness} introduce a loss term directly in the model training objective to make the explanation more robust instead. For an input $\x$, the model output's logit $f(\x)$ and the remaining logits $f(\x)_{n}$ where $n = 1 \dots N$ in $N$-class classification, consider the gradient explanation $\E_{\hat{y}} = \E(f(\x), \x) = \nabla_{x}(f(\x))$. For the $j^{\text{th}}$ class where $j = \argmax_{j \neq \hat{y}} f(x)_{j}$, consider the explanation $\E_{j} = \E(f(\x)_{j}, \x) = \nabla_{x}(f(\x)_{j})$. With the intuition that the gradient of the output logit $f(\x)$ should be least dissimilar to the input $\x$ and the gradient of other logits should be most dissimilar, the metric \textbf{robustness-loss} can be formalized for a perturbed input $\x'$ as:

\begin{equation*}
\text{robustness-loss}(\E, f, \x') \defines \log{(1 + \exp{(- (\text{cosine-dissimilarity($\E_{j} , \x'$)} - \text{cosine-dissimilarity($\E_{\hat{y}},\x'$)})}}
\end{equation*}
\begin{flalign}\label{robustness_loss}
&& \triangleright\,\text{Equation 3 in \citeauthor{Singh2019AttributionalRobustness}}
\end{flalign}

The complete loss function can then be formalized as the sum of two terms for the worst-case perturbed input $\x'$ -- $\mathcal{L}\left(\E, f, \x'\right)$ denoting the classification loss and $\text{robustness-loss}(\E, f, \x')$. The worst-case perturbation can be obtained as $\x' = \argmax_{\ell_p(\x'-\x) \leq r} \text{robustness-loss}(\E, f, \x')$. By minimizing the joint loss term computed for the worst-case perturbation, one can train the model to generate robust explanations within a local neighborhood of $\x$.

\begin{flalign}\label{accuracy_robustness_loss}
&&\text{accuracy-and-robustness-loss}(\E, f, \x') \defines \mathcal{L}\left(\E, f, \x'\right) + \lambda \cdot \text{robustness-loss}(\E, f, \x') && \triangleright\,\text{Equation 2 in \citeauthor{Singh2019AttributionalRobustness}}
\end{flalign}

\citet{Chen2019Regularization} repurpose integrated gradients \citet{AxiomaticAttribution} for intermediate model representations (layers). Let $R = [R_1, R_2, \dots R_n]$ be a hidden layer in the model $f$ comprising $n$ neurons. Then $R(\x)$ is the function induced by the previous layers. Let the curve $c : [0, 1] \rightarrow \mathbb{R}^K$ trace the movement from $\x$ to $\x'$, such that $c(0) = \x$ and $c(1) = \x'$. The composition $R \circ c$ represents a new curve in the hidden layer space, tracing the movement from $R(\x)$ to $R(\x')$. With a differentiable model loss $\mathcal{L}$, the attribution for a neuron $R_{i}$ can be formulated as: 

\begin{flalign}\label{integrated_gradients}
    &&\text{integrated-gradients}_{R_{i}}(\mathcal{L}, \x, \x') \defines \sum_{j=1}^{K} \int_{0}^1 \frac{\partial \mathcal{L}(R(c(t))}{\partial R_{i}} \frac{\partial R_{i} (c(t))}{\partial \x_{j}} c_{j}'(t) dt && \triangleright\,\text{Equation 2 in \citeauthor{Chen2019Regularization}}
\end{flalign}

Further,\citet{Chen2019Regularization} jointly penalize the model loss $\mathcal{L}$ and the attribution of changes to $\mathcal{L}$ when we move from $\x$ to $\x'$, via \textbf{integrated-gradient-regularization}. Let $p$-norm $\ell_p$ to be the size function of $\text{integrated-gradients}_{R}$.

\begin{equation*}
\text{integrated-gradient-regularization}(f, \x') \defines \mathcal{L}(y, f(\x)) + \lambda \max_{\left\Vert \x'-\x \right\Vert \leq r} \ell_p(\text{integrated-gradients}_R(\mathcal{L}, \x, \x'))
\end{equation*}
\begin{flalign}\label{integrated_gradient_regularization}
&& \triangleright\,\text{Equation 4 in \citeauthor{Chen2019Regularization}}
\end{flalign}

\end{document}